%% file: main.tex
\documentclass{article}
\usepackage{ijcai21}

\input{math_commands.tex}

\usepackage{times}
\usepackage{soul}
\usepackage{url}
\usepackage[hidelinks]{hyperref}
\usepackage[utf8]{inputenc}
\usepackage[small]{caption}
\usepackage{graphicx}
\usepackage{amsmath}
\usepackage{amsthm}
\usepackage{booktabs}
\usepackage{algorithm}
\usepackage{algorithmic}
\usepackage{float}
\usepackage[T1]{fontenc}
\usepackage{bm}
\usepackage{subfigure}
\usepackage{wrapfig}
\usepackage{helvet} 
\usepackage{courier}  
\usepackage{graphicx} 
\usepackage{caption} 
\usepackage{amssymb}
\usepackage{latexsym}
\usepackage{bbding}
\usepackage{booktabs}
\usepackage{multirow}
\usepackage{color}
\usepackage{authblk}

\newtheorem{theorem}{\bf Theorem}
\newtheorem{lemma}{\bf Lemma}
\newtheorem{proposition}{\bf Proposition}
\newtheorem{corollary}{\bf Corollary}
\newtheorem{definition}{\bf Definition}

\newtheorem{assumption}{\bf Assumption}

\title{Certified Distributional Robustness on Smoothed Classifiers}
\author[$ \dagger $]{Jungang~Yang}
\author[$ \dagger $]{Liyao~Xiang \thanks{Liyao Xiang (xiangliyao08@sjtu.edu.cn) is the corresponding author with the John Hopcroft Center, Shanghai Jiao Tong University, China.}}
\author[$ \dagger $]{Ruidong~Chen}
\author[$ \dagger $]{Yukun~Wang}
\author[$ \ddagger $]{Wei~Wang}
\author[$ \dagger $]{Xinbing~Wang}
\affil[$ \dagger $]{Shanghai Jiao Tong University}
\affil[$ \ddagger $]{Hong Kong University of Science and Technology}

\begin{document}

\maketitle

\begin{abstract}

The robustness of deep neural networks (DNNs) against adversarial example attacks has raised wide attention. For smoothed classifiers, we propose the worst-case adversarial loss over input distributions as a robustness certificate. Compared with previous certificates, our certificate better describes the empirical performance of the smoothed classifiers. By exploiting duality and the smoothness property, we provide an easy-to-compute upper bound as a surrogate for the certificate. We adopt a noisy adversarial learning procedure to minimize the surrogate loss to improve model robustness. We show that our training method provides a theoretically tighter bound over the distributional robust base classifiers. Experiments on a variety of datasets further demonstrate superior robustness performance of our method over the state-of-the-art certified or heuristic methods.

\end{abstract}

\input{intro}

\input{related}

\input{formulation}

\input{comparison}

\input{experiment}
\input{result}

\input{conclusion}

\bibliographystyle{named}
\bibliography{main}

\appendix
\input{appendix}

\input{appendix_B}

\end{document}

%% file: math_commands.tex
\usepackage{amsmath,amsfonts,bm}

\def\eqref#1{equation~\ref{#1}}

\def\1{\bm{1}}

\def\mI{{\bm{I}}}

\DeclareMathAlphabet{\mathsfit}{\encodingdefault}{\sfdefault}{m}{sl}
\SetMathAlphabet{\mathsfit}{bold}{\encodingdefault}{\sfdefault}{bx}{n}

\DeclareMathOperator*{\argmax}{arg\,max}

%% file: intro.tex
\section{Introduction}

Deep neural networks (DNNs) have been known to be vulnerable to adversarial example attacks: by feeding the DNN with slightly perturbed inputs, the attack alters the prediction output. The attack can be fatal in performance-critical systems such as autonomous vehicles. A classifier is {\em robust} when it can resist such an attack that, as long as the range of the perturbation is not too large (usually invisible by human), the classifier produces an expected output despite of the specific perturbation. A {\em certifiably robust} classifier is one whose prediction at any point $x$ is verifiably constant within some set around $x$. 

A conventional way to obtain a certifiable robust classifier is to perform randomized smoothing \cite{cohen2019certified,pinot2019theoretical,li2019certified,lecuyer2019certified}. Assume a base classifier $f$ tries to map instance $x_0$ to corresponding label $y$. It is found that when fed with instance $x$ perturbed from $x_0$, the smoothed classifier $g(x) = \mathbb{E}_{z}[f(x+z)]$ provably returns the same label as $g(x_0)$ does.

We observe that, even $g(x)$ does not alter from $g(x_0)$ under adversarial perturbations, there is no guarantee that $g(x_0)$ would return $y$. It is possible that the adversarially perturbed input has the same label as the original one which is wrongly classified by $g$. In fact, previous certificates are derived instance-wise, meaning that for one particular instance, there is a distortion bound within which the prediction result does not vary. However, it is unknown how the smoothed classifier performs on the input sample population, which we think is an important robustness indicator since it directly relates to the empirical accuracy of the classifier. 

We propose a distributional robustness certificate for smoothed classifiers. We postulate the inputs to the classifier are drawn from a data-generating distribution, and there is a class of distributions around the data-generating distribution. The problem is to seek one that maximizes the loss over a smoothed classifier. It is clear that such a certificate is associated with the empirical accuracy: under the same amount of perturbation, a classifier with a smaller worst-case adversarial loss enjoys higher accuracy over the input population. We prove that the smoothed classifier $g(\cdot)$ typically has a tighter certificate than its corresponding base classifier $f(\cdot)$, suggesting higher robustness against adversarial examples.

To obtain a robust DNN, we minimize the above worst-case adversarial loss over the input distribution. Let $\ell(\cdot)$ be the loss function and the classifier be parameterized by $\theta \in \Theta$. The perturbed input is $s = x+z \sim P \oplus Z $ where $ P \oplus Z $ is the joint distribution of adversarial examples and the Gaussian noise. We aim at minimizing the following certificate:
\begin{equation}
\mathop{\text{minimize}}_{\theta \in \Theta} \sup_{P \oplus Z}\mathbb{E}_{P \oplus Z}[\ell(\theta; s)].
\end{equation}

However, it still remains problematic how to obtain the above distributional robustness certificate in practice. Therefore, we derive an upper bound for the worst-case adversarial loss of smoothed classifiers and optimize the upper bound as a surrogate loss. We show that the optimal surrogate loss is still inferior to the worst-case adversarial loss of the base classifier, and can be obtained by a noisy adversarial learning procedure. From the training perspective, our method can be considered as augmenting the adversarial examples with Gaussian noise so that instead of minimizing over a limited number of adversarial examples, we minimize over a larger adversarial region where the expected perturbation loss is the worst. As a result, the trained classifier is more robust since it has seen a well-depicted adversarial example distribution.

Compared to vanilla adversarial training \cite{madry2017towards,phan2019heterogeneous}, our approach does not seek one single data point, but rather a neighborhood around the adversarial examples that maximizes the loss in the inner maximization. Compared to adversarial distributional training (ADT) \cite{deng2020adversarial}, our method models the worst-case changes in data distribution rather than a worst-case adversarial distribution around the natural inputs. Moreover, we provide provable robustness guarantee for our training method. Other training methods with a robustness certificate, such as distributionally robust optimization \cite{ben2013robust,esfahani2018data,sinha2017certifying}, seek a data-generating distribution which generates the worst-case adversarial examples, while we are modeling a data-generating distribution which is jointly composed by the worst-case adversarial distributions of the dataset. Compared to previous robustness certificates via smoothed classifiers \cite{lecuyer2019certified,li2019certified,cohen2019certified,lee2019tight,salman2019provably}, our method provides a provable guarantee w.r.t. the ground truth input distribution, which better illustrates the robustness of a DNN than the distortion range-based certificate.

Highlights of our contribution are as follows. {\em First,} we propose a distributional robustness certificate over noisy inputs, and such a certificate better captures the empirical performance for the smoothed classifiers. {\em Second,} we prove the advantage of smoothed classifiers over the base ones from the perspective of certificates, implying a more robust performance for the deployed models. {\em Third,} we derive a data-dependent upper bound for the certificate, and minimize it in the training loop. The smoothness property entails the computational tractability of the certificate. We conduct extensive experiments on MNIST, CIFAR-10 and TinyImageNet, comparing with the state-of-the-art adversarial training methods as well as randomized smoothing based methods. The experimental results demonstrate that our method excels in empirical robustness.


%% file: related.tex
\section{Related Work}
Works proposed to defend against adversarial example attacks can be categorized as follows.

In \textbf{empirical defences,} there is no guarantee how the DNN model would perform against the adversarial examples. Stability training (\cite{zheng2016improving,zantedeschi2017efficient}) improves model robustness by adding randomized noise to the input during training but shows limited performance enhancement. Adversarial training (\cite{kurakin2016adversarial,madry2017towards,zhang2019theoretically,wang2019improving}) trains over adversarial examples found at each training step but unfortunately does not guarantee the performance over unseen adversarial inputs. Although without a guarantee, adversarial training has excellent performance in empirical defences against adversarial attacks.

\textbf{Certified defences} are certifiably robust against any adversarial input within an $\ell_{p}$-norm perturbation range from the original input. A line of works construct computationally tractable relaxations for computing an upper bound on the worst-case loss. The relaxations include linear programming (\cite{wong2018provable}), mixed integer programming (\cite{tjeng2018evaluating}), semidefinite programming (\cite{raghunathan2018semidefinite}), and convex relaxation (\cite{namkoong2017variance,salman2019convex}). 
\cite{sinha2017certifying} also propose a robustness certificate based on a Lagrangian relaxation of the loss function, and it is provably robust against adversarial inputs drawn from a distribution centered around the original input distribution. The certificate of our work is constructed on a Lagrangian relaxation form of the worst-case loss, but has a broader applicability than \cite{sinha2017certifying} with a tighter loss bound due to the smoothness property.


\textbf{Randomized smoothing} introduces randomized noise to the neural network, and tries to provide a statistically certified robustness guarantee. 
The smoothing method does not depend on a specific neural network, or a type of relaxation, but can be generally applied to arbitrary neural networks. The idea of adding randomized noise was first proposed by \cite{lecuyer2019certified}, given the inspiration of the differential privacy property, and then \cite{li2019certified} improve the certificate with R\'{e}nyi divergence. \cite{cohen2019certified} obtain a larger certified robustness bound through the smoothed classifier based on Neyman-Pearson theorem. \cite{phan2019scalable} extend the noise addition mechanism to large-scale parallel algorithms. By extending the randomized noise to the general family of exponential distributions, \cite{pinot2019theoretical} unify previous approaches to preserve robustness to adversarial attacks. \cite{lee2019tight} offer adversarial robustness guarantees for $ \ell_0 $-norm attacks. Both \cite{salman2019provably,jia2019certified} employ adversarial training to improve the performance of randomized smoothing. Following a similar principle, our work trains over adversarial data with randomized noise. But we provide a more practical robustness certificate and a training method achieving higher empirical accuracy than theirs.

%% file: formulation.tex
\section{Proposed Approach}
We first define the closeness between distributions, based on which we depict how far the input distribution is perturbed. Under the perturbation constraint, we introduce the robustness certificate for smoothed classifiers. Our main theorem gives a tractable robustness certificate which is easy to optimize. Following the certificate, we illustrate our algorithm for improving the robustness of the smoothed classifiers. All proofs are collected in the appendices for conciseness.

\subsection{A Distributional Robustness Certificate}
\begin{definition}[Wasserstein distance]\label{def:Wasserstein}
	Wasserstein distances define a notion of closeness between distributions. Let $\left(\mathcal{X} \subset \mathbb{R}^{d}, \mathcal{A}, P\right)$ be a probability space and the transportation cost $c: \mathcal{X} \times \mathcal{X} \rightarrow[0, \infty)$ be nonnegative, lower semi-continuous, and $c(x, x)=0.$ $P$ and $Q$ are two probability measures supported on $\mathcal{X}$. Let $\Pi(P, Q)$ denotes the collection of all measures on $ \mathcal{X} \times \mathcal{X} $ with marginals $P$ and $Q$ on the first and second factors respectively, {\em i.e.,} it holds that $\pi(A, \mathcal{X})=P(A)$ and $\pi(\mathcal{X}, A)=Q(A), ~\forall A\in \mathcal{A}~ \text{and} ~\pi \in \Pi(P,Q).$ The Wasserstein distance between $P$ and $Q$ is
\begin{equation}
W_{c}(P, Q):=\inf _{\pi \in \Pi(P, Q)} \mathbb{E}_{\pi}\left[c\left(x, y\right)\right].
\end{equation}
\end{definition}
For example, the $ \ell_{2} $-norm $ c(x,x_0) = \|x-x_0\|^2_2 $ satisfies the aforementioned conditions.

\textbf{Distributional robustness for smoothed classifiers.} Assume the original input $x_{0}$ is drawn from the distribution $P_{0}$, and the perturbed input $x$ is drawn from the distribution $P$. Since the perturbed input should be visually indistinguishable from the original one, we define the robustness region for the smoothed classifier as ${\mathcal{P}}=\left\{{P}: W_{c}\left(P\oplus Z, P_{0}\right) \leq \rho, {P}\in P(\mathcal{X})\right\}$ where $\rho > 0$.  Instead of regarding the noise as a part of the smoothed classifier, we let $s = x + z$ be a noisy input coming from the distribution $P\oplus Z$. We use $ p_{x}, p_{z} $ to denote the probability density function of $ x, z. $ The probability density function of $ s $ can be written as:
\begin{equation}
p_{s}(s) = \int_{\mathbb{R}^d} p_{x}(t) p_{z}(s-t) \,dt.
\end{equation}
Since the noise $z \in \mathbb{R}^{d}$, we need to set $ \mathcal{X} = \mathbb{R}^d $ to admit $ x+z \in \mathcal{X} $ as \cite{lecuyer2019certified,cohen2019certified,salman2019provably} do.  Within such a region, we evaluate the robustness as a worst-case population loss over noisy inputs: $\sup_{P: W_c(P \oplus Z, P_0) \le \rho } \mathbb{E}_{P \oplus Z}  [\ell(\theta ; s)]$. Essentially, we evaluate the robustness of the smoothed classifer based on its performance on the worst-case adversarial example distribution. It is easy to show that the worst-case adversarial loss is smaller than the one without noise:

\begin{theorem}\label{thm:smoothing}
Let  $\ell: \Theta \times \mathcal{X} \rightarrow \mathbb{R}$, $ x_0$ be an input drawn from the input distribution $P_0$, $ x $ be the adversarial example which follows the distribution $ P $ and  $ z \sim Z = \mathcal{N}(0, \sigma^2I) $ be the additive noise of the same shape as $ x $. The sum of $x$ and $z$ is denoted as $ s = x+z \sim P \oplus Z $, we have
\begin{equation}\label{smoothing inequation}
\begin{split}
\sup _{P: W_c(P \oplus Z, P_0) \le \rho } \mathbb{E}_{P \oplus Z}  [\ell(\theta ; s )] \le   \sup _{ P^{\prime}: W_{c}(P^{\prime}, P_0) \le \rho} \mathbb{E}_{P^{\prime}}[\ell(\theta ; x')].
\end{split}
\end{equation}
\end{theorem}

The proof is straight-forward. The right-hand side of Eq.~\ref{smoothing inequation} indicates the worst-case loss over all adversarial distributions which are at most $\rho$ away from $P_{0}$. It is intuitive that the one with Gaussian noise is a special case of the adversarial distributions and hence its worst-case loss should be no larger. The theorem illustrates that within a given perturbation range, the smoothed classifier potentially provides a lower adversarial loss over the input distribution and therefore higher robustness. However, such a robustness metric is impossible to measure in practice as we cannot depict $P$ precisely. Even if $P$ can be acquired, it can be a non-convex region which renders the constrained optimization objective intractable. Hence we resort to the Lagrangian relaxation of the problem to derive an upper bound for it.

\subsection{A Surrogate Loss}
As the main theorem of this work, we provide an upper bound for the worst-case population loss for any level of robustness $\rho$. We further show that for small enough $\rho$, the upper bound is tractable and easy to optimize.
\begin{theorem}\label{thm:robust_bound}
	Let $\ell: \Theta \times \mathcal{X} \rightarrow \mathbb{R}$ and transportation cost function $c: \mathcal{X} \times \mathcal{X} \rightarrow \mathbb{R}_{+}$ be continuous. Let $ x_0$ be an input drawn from the input distribution $P_0$, $ x $ be the adversarial example which follows the distribution $ P $ and  $ z \sim Z = \mathcal{N}(0, \sigma^2I) $ be the additive noise of the same shape as $ x $. We let $\phi_{\gamma}\left(\theta ; x_{0}\right)= \sup _{x \in \mathcal{X}} \mathbb{E}_{Z} \left\{\ell(\theta ; x+z)-\gamma c\left(x+z, x_{0}\right)\right\}$ be the robust surrogate. For any $\gamma, \rho>0$ and $ \sigma $, we have
	\begin{equation}\label{eq:thm1_1}
	\sup _{ P: W_c(P \oplus Z, P_0) \le \rho } \mathbb{E}_{P \oplus Z} [\ell(\theta ; s)] \le \gamma \rho+\mathbb{E}_{P_0} \left[\phi_{\gamma}(\theta ; x_0)\right].
	\end{equation}
\end{theorem}

The proof is given in supplementary file A.1. It is notable that the right-hand side take the expectation over $P_0$ and $Z$ respectively. Given a particular input $x_{0} \sim P_0$, we seek an adversarial example that maximizes the expected loss over the additive noise. Typically, $P_0$ is impossible to obtain and thus we use an empirical distribution, such as the training data distribution, to approximate $P_0$ in practice. 

Since Thm.~\ref{thm:robust_bound} provides an upper bound for the worst-case population loss, it offers a principled adversarial training approach which minimizes the upper bound instead of the actual loss, {\em i.e.,} 
\begin{equation}\label{our loss function}
\underset{\theta \in \Theta}{ \rm minimize} ~~\mathbb{E}_{P_0} [\phi_{\gamma}(\theta ; x_0)].
\end{equation}
In the following we show the above loss function has a tractable form for arbitrary neural networks, due to a smoothed loss function. Hence Thm.~\ref{thm:robust_bound} provides a tractable robustness certificate depending on the data.

\textbf{Properties of the smoothed classifier.} We show the optimization objective of Eq.~\ref{our loss function} is easy to compute for any neural network, particular for the non-smooth ones with ReLU activation layers. More importantly, the smoothness of the classifier enables the adversarial training procedure to converge as we want by using the common optimization techniques such as stochastic gradient descent. The smoothness of the loss function comes from the smoothed classifier with randomized noise. Specifically,
\begin{lemma}\label{thm:smooth}
	Assume $ \ell : \Theta \times\mathcal{X} \rightarrow [0,M] $ is a bounded loss function. The loss function on the smoothed classifier can be expressed as $\hat{\ell}(\theta; x) :=  \mathbb{E}_{Z} [\ell(\theta ; x+z)],~z \sim Z = \mathcal{N}(0,\sigma^2I)$. Then we have $ \hat{\ell} $ is $ \frac{2M}{\sigma^2} $-smooth w.r.t. $ \ell_{2} $-norm, {\em i.e.,} $ \hat{\ell} $ satisfies
	\begin{equation}
	\left\|\nabla_{x} \hat{\ell}(\theta ; x)-\nabla_{x} \hat{\ell}\left(\theta ; x^{\prime}\right)\right\|_{2} \leq \frac{2M}{\sigma^2} \left\|x-x^{\prime}\right\|_2 .
	\end{equation}
\end{lemma}
The proof is in supplementary file A.2. It mainly takes advantage of the randomized noise which has a smoothing effect on the loss function. For DNNs with non-smooth layers, the smoothed classifier makes it up and turns the loss function to a smoothed one, which contributes as an important property to the strong concavity of $\mathbb{E}_{Z} \left[ \ell(\theta ; x+z)-\gamma c\left(x+z, x_{0}\right)\right]$ and therefore ensures the tractability of the robustness certificate. Please refer to supplementary file A.3 and A.6 for more properties.

\subsection{Noisy Adversarial Learning Alogrithm}
Problem~(\ref{our loss function}) provides an explicit way to improve the robustness of a smoothed classifier parameterized by $\theta$. We correspondingly design a noisy adversarial learning algorithm to obtain a classifier of which its robustness can be guaranteed. In the algorithm, we use the empirical distribution to replace the ideal input distribution $P_0$, and sample $z$~ $r$ times to substitute the expectation with the sample average. Assuming we have a total of $n$ training instances $x_{0}^{i}, \forall i\in [n]$, and sample $z_{ij} \sim \mathcal{N}(0, \sigma^2 I)$ for the $i$-th instance for $r$ times, the objective is:
\begin{equation}\label{sample loss function}
\underset{\theta \in \Theta}{ \rm minimize} ~~\frac{1}{nr}\sum_{i=1}^{n}\sup _{x \in \mathcal{X}} \sum_{j=1}^{r} \left[\ell(\theta ; x+z_{ij})-\gamma c\left(x+z_{ij}, x_{0}^{i}\right)\right].
\end{equation}
The detail of the algorithm is illustrated in Alg.~\ref{alg:NAL}. In the inner maximization step (line 3-6), we adopt the \textit{projected gradient descent} (PGD \cite{madry2017towards,kurakin2016adversarial}) to approximate the maximizer according to the convention. The hyperparameters include the number of iterations $K$ and the learning rate $ \eta_1 $. Within each iteration, we sample the Gaussian noise $r$ times, given which we compute an average perturbation direction for each update. The more noise samples, the closer the averaging result is to the expected value, which is at the sacrifice of higher computation expense. Similarly, a larger number of $K$ indicates stronger adversarial attacks and higher model robustness, but also incurs higher computation complexity. Hence choosing appropriate values of $r$ and $K$ is important in practice.

\begin{algorithm}
	\renewcommand{\algorithmicrequire}{\textbf{Input:}}
	\renewcommand{\algorithmicensure}{\textbf{Output:}}
	\caption{Training Phase of NAL}
	\label{alg:NAL}
	\begin{algorithmic}[1] 
		\REQUIRE number of training samples $n$, number of noise samples $r$, noise STD $ \sigma $, learning rate $ \eta_1,\eta_2 $, number of iterations $K$, penalty parameter $ \gamma  $, training iterations $T$
		\ENSURE the classifier parameter $ \theta $
		\FOR {$ t \in \{1,\ldots,T\} $}
		\FOR {$i \in \{1,\ldots,n\}$}
		\FOR {$ k \in \{0,\ldots,K-1\} $}
		\STATE{$ \Delta x_{k}^{i} = \frac{1}{r} \sum_{j=1}^{r} \bigtriangledown_{x_{k}^{i}} \ell(\theta;x_{k}^{i}+z_{ij}) - \gamma \bigtriangledown_{x_{k}^{i}} c(x_{k}^{i}+z_{ij}, x_{0}^{i}) $, where $ z_{ij} \sim \mathcal{N}(0,\sigma^2I) $}
		\STATE{$ x_{k+1}^{i} = x_{k}^{i} + \eta_1 \Delta {x_{k}^{i}} $}
		\ENDFOR
		\ENDFOR
		\STATE {$ \theta^{t+1} = \theta^{t} - \eta_2 \left\{ \frac{1}{nr} \sum_{i=1}^{n} \left[ \bigtriangledown_{\theta} \sum_{j=1}^{r} \ell(\theta^{t}; x_{K}^{i} + z_{ij} ) \right]  \right\} $}
		\ENDFOR
	\end{algorithmic} 
\end{algorithm}

After training is done, we obtain the classifier parameter $\theta$. In the inference phase, we sample a number of $z \sim \mathcal{N}(0, \sigma^2 I)$ to add to the testing instance. The noisy testing examples are fed to the classifier to get the prediction outputs.

{\bf An alternative view.} We provide an alternative intuition for our algorithm. Assume that the input dataset contains $n$ records and $ X_0 $ is a sample of the distribution $ P_0^{n} $ since each input $x_0 \sim P_0.$  Correspondingly, let the adversarial sample set be $X$ which is drawn from the worst-case adversarial distribution $P^{\star n},~\text{and}~ P^{\star} = {\arg} \sup_{P \in \mathcal{P} } \mathbb{E}_{P}  [\ell(\theta ; x)] $. A significant drawback of the vanilla adversarial training is that, the loss value at each training sample could be quite different, resulting in high instability in the model robustness. Instead of training over a limited number of inputs, we wish to train over a larger input set to reduce the instability. Ideally, we train over $m$ samples of $P^{\star n}$ rather than one: $X^{(1)}, \ldots ,X^{(m)}\sim P^{\star n}.$ Obviously, we do not have so much data, so we adopt a sampled mean $ M = {\sum_{i=1}^{m}X^{(i)}}/{m} $ for training. It is intuitively more stable to train over $ M $ than over $X$. To acquire $M$, we apply \textit{Central Limit Theorem} for an estimation: as $ m\rightarrow \infty $, $M \sim \mathcal{N}(\mu, \frac{\varrho^2}{m}I),$ where $ \mu $ and $ \varrho^2 $ are the mean and variance of the distribution $ P^{\star n} $. By using the worst-case adversarial examples $ X $ to approximate $ \mu $, and $ \sigma^2 $ to  estimate $ \frac{\varrho^2}{m},$ our algorithm trains over $\hat{M} \sim \mathcal{N}(X, \sigma^2I).$ It means that each input is added the Gaussian noise $z \sim Z = \mathcal{N}(0, \sigma^2 I)$ before being fed to the classifier. The trick is called {\em randomized smoothing} in \cite{lecuyer2019certified,cohen2019certified,salman2019provably}. Hence, we show an alternative explanation why randomized smoothing could enhance the robustness of the classifiers.

\textbf{Convergence.} An important property associated with the smoothed classifier is the strong concavity of the robust surrogate loss, which is the key to the convergence proof. 
The detail of the proof can be found in supplementary file A.4. 
As long as the loss $ \hat{\ell} $ is smooth on the parameter space $ \Theta $, NAL has a convergence rate $O(1/\sqrt{T})$, similar to \cite{sinha2017certifying}, but NAL does not need to replace the non-smooth layer ReLU with Sigmoid or ELU to guarantee robustness.

%% file: comparison.tex
\begin{figure*}[htbp]
	\centering
	\subfigure[]{
		\begin{minipage}[t]{0.245\linewidth}
			\centering
			\includegraphics[width=1\linewidth]{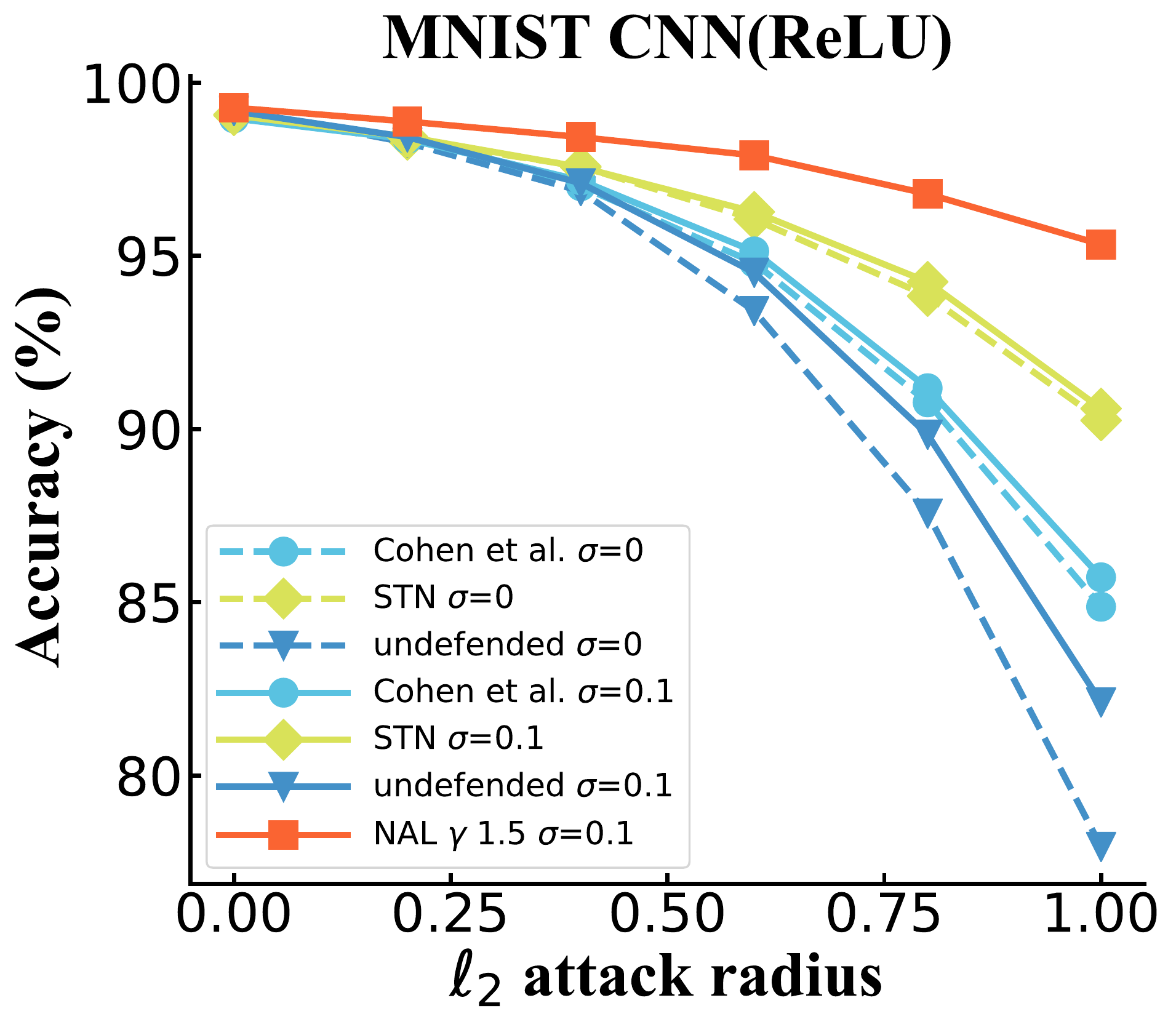}
		\end{minipage}%
	}%
	\subfigure[]{
		\begin{minipage}[t]{0.245\linewidth}
			\centering
			\includegraphics[width=0.95\linewidth]{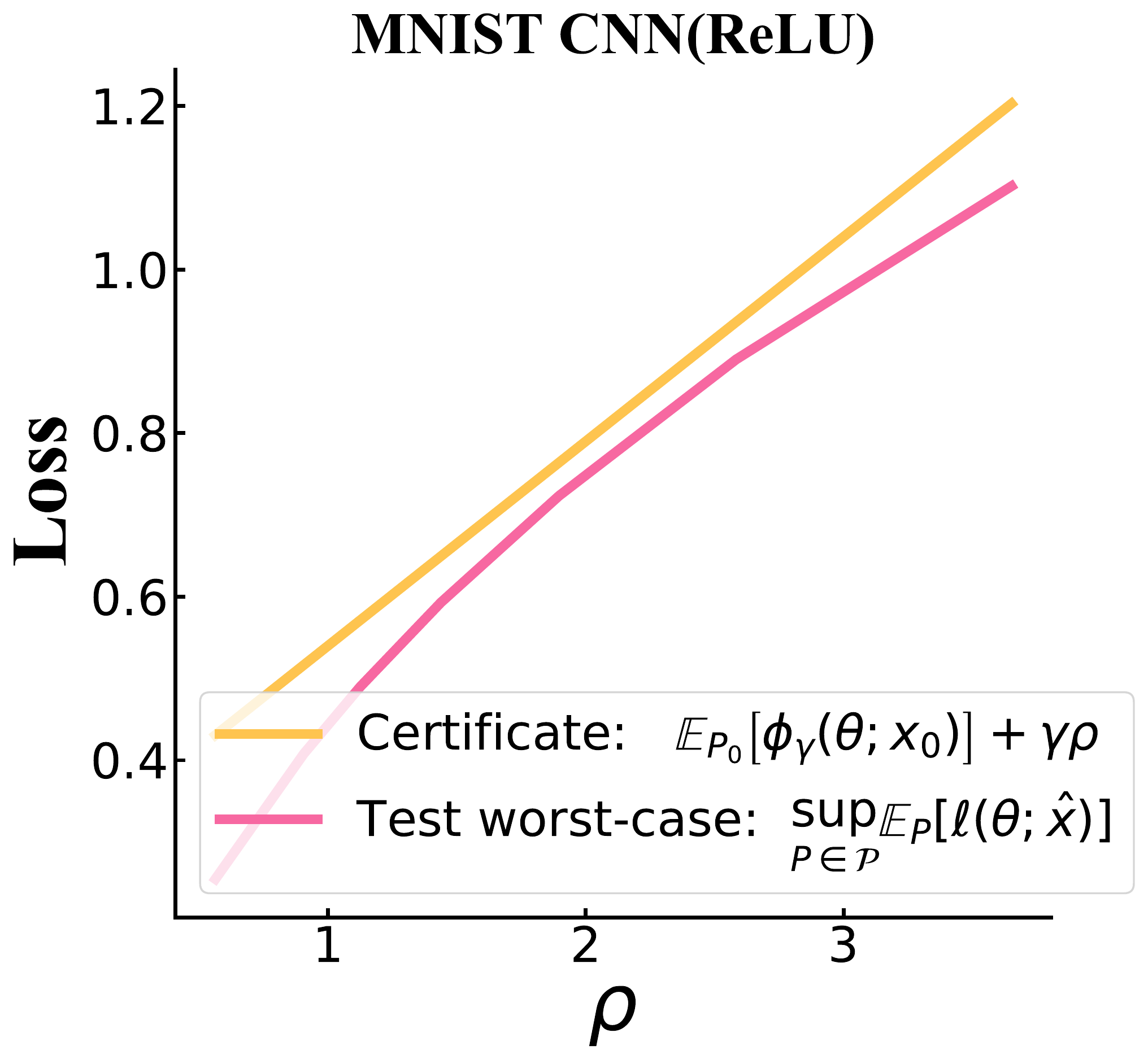}
		\end{minipage}%
	}%
	\subfigure[]{
		\begin{minipage}[t]{0.265\linewidth}
			\centering
			\includegraphics[width=1\linewidth]{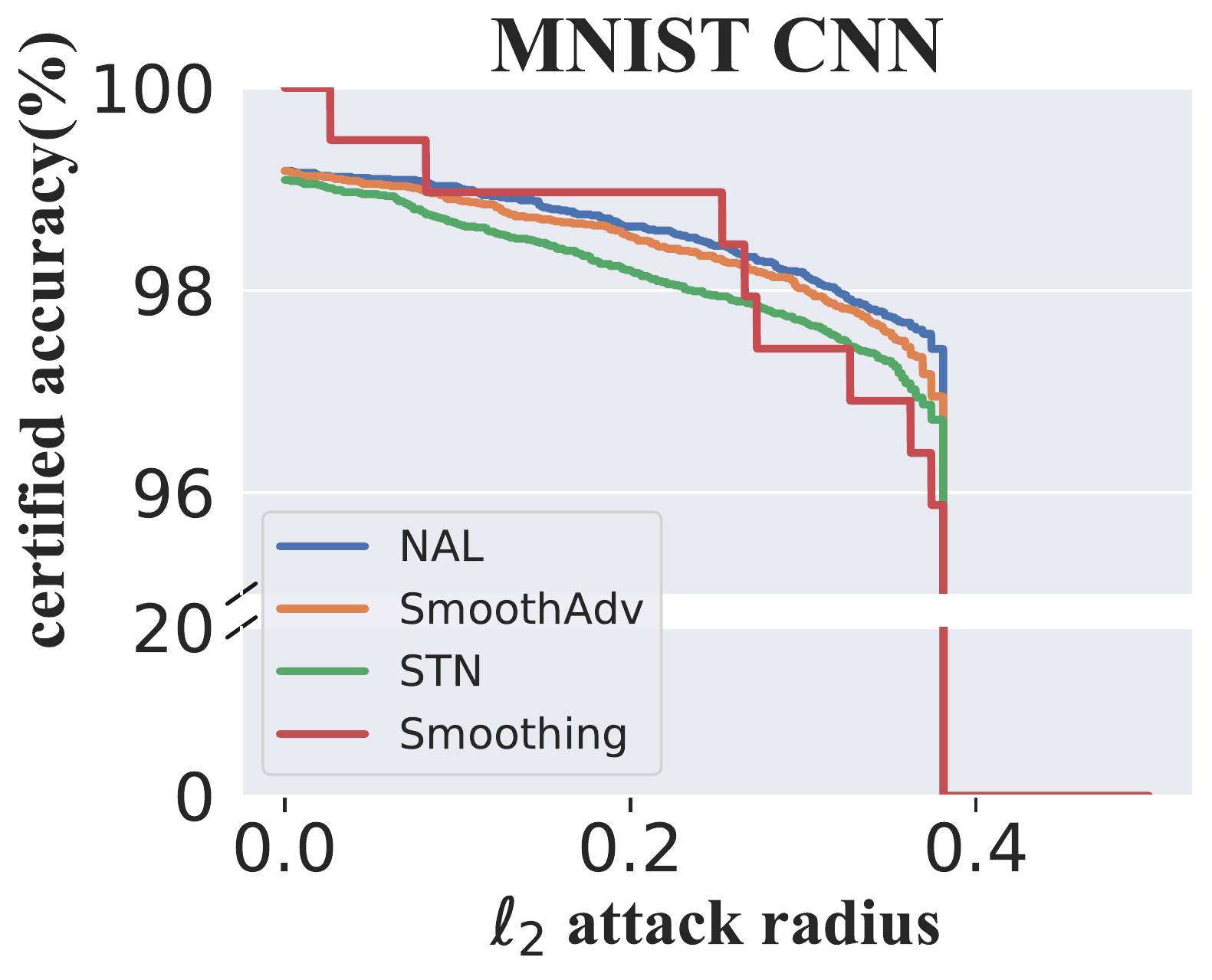}
			
		\end{minipage}%
	}%
	\subfigure[]{
		\begin{minipage}[t]{0.245\linewidth}
			\centering
			\includegraphics[width=1\linewidth]{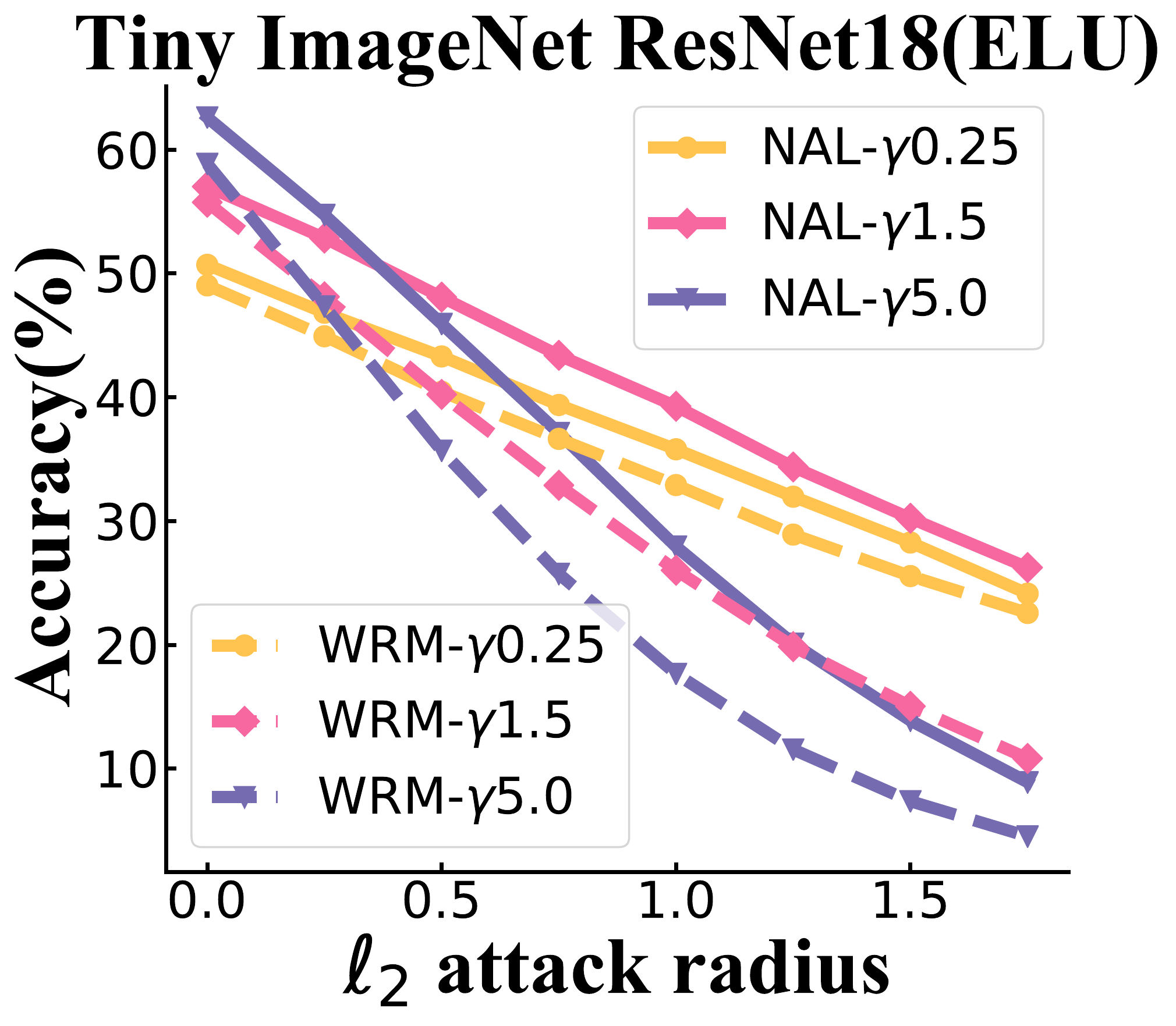}
		\end{minipage}%
	}%
	\vspace{-4mm}
	\caption{ (a) Accuracies of models trained on MNIST under different levels of $\ell_2$ attacks. Undefended means a naturally trained model. Solid lines represent models tested with additive noise, and dotted lines mean models are tested without noise. $ \sigma = 0.1 $ represents adding Gaussian noise $ \mathcal{N}(0, 0.1^2 \mI) $ to the testing samples. (b) gives the distance between the robustness certificate (yellow) and the worst-case performance on testing data (pink) with an example on MNIST. The gap between the two lines indicates the tightness of our certificate (Eq.~\ref{eq:thm1_1}). (c) compares NAL with SmoothAdv, Smoothing and STN on MNIST at $ \gamma=1.5 $ and the corresponding $ \varepsilon $. NAL shows a close certified accuracy to SmoothAdv while superior to other baselines overall. (d) NAL outperforms WRM on Tiny ImageNet, ResNet-18 (ELU) under different $ \gamma $s. }
	\label{fig:certify}
	\vspace{-5mm}
\end{figure*}

\section{Comparison with Other Certificates }
We compare our work with the state-of-the-art robustness definitions and certificates in this section.

\subsection{Adversarial Training}
In Thm. \ref{thm:smoothing}, we show that the distributional robustness certificate improves over smoothed classifiers. Now we further demonstrate that our robust surrogate loss is still inferior to the worst-case adversarial loss of the base classifier. Note that the robustness certificate of the base classifier is given by \cite{sinha2017certifying}.

\begin{corollary}\label{thm: tighter bound}
	Under the same denotations and conditions as Thm.~\ref{thm:robust_bound}, we have
	\begin{equation}\label{tighter bound}
   	\resizebox{.91\linewidth}{!}{$
		\displaystyle
	\begin{split}
	\inf_{\gamma \geq 0} \left\{\gamma \rho+\mathbb{E}_{P_0}\left[\phi_{\gamma}(\theta ; x_0)\right]\right\}
	\leq  \sup _{ P^{\prime}: W_{c}(P^{\prime}, P_0) \le \rho} \mathbb{E}_{P^{\prime}}[\ell(\theta ; x')].
	\end{split}
	$}
	\end{equation}
\end{corollary}
The proof is given in supplementary file A.5. We demonstrate that for the smoothed classifier, not only the worst-case loss could be smaller, but the proposed upper bound is no larger than the certificate of the base classifier. If the outer minimization problem applies to both sides of the inequality, our approach would potentially obtain a smaller loss when both classifiers share the same neural architecture.

\subsection{Smoothed Classifiers}
\label{sec: smoothed classifiers}
Works including \cite{lecuyer2019certified,cohen2019certified,pinot2019theoretical,li2019certified} and others guarantee the robustness of a DNN classifier by inserting randomized noise to the input at the inference phase. Most of them do not concern about the training phase, but merely illustrate the relation between robustness certificate and the magnitude of the noise. For example, assume the original input $x_{0}$ and its perturbation $x$ are within a given range $\left\| x - x_{0} \right\|_{2} \leq \varepsilon.$ The smoothed classifier $g(x)$ returns class $c_i$ with probability $p_i$. For instance $x_0$, robustness is defined by the largest perturbation radius $R$ which does not alter the instance's prediction, {\em i.e.,} $g(x)$ is classified into the same category as $g(x_0)$. Such perturbation radius depends on the largest and second largest probabilities of $p_i$, denoted by $ p_A, p_B $ respectively. For example, the results in \cite{cohen2019certified} have shown that $R=\frac{\sigma}{2}\left(\Phi^{-1}\left(\underline{p_{A}}\right)-\Phi^{-1}\left( \overline{p_{B}}\right)\right)$ where $\Phi^{-1}$ is the inverse of the standard Gaussian CDF, $\underline{p_{A}}$ is a lower bound of $p_{A}$, and $\overline{p_{B}}$ is an upper bound of $p_{B}$.

The previous robustness definition only guarantees $g(x)$ to be classified to the same class as $g(x_0)$, but ignores the fact that $g(x_0)$ may be wrongly classified, especially when $x_0$ is perturbed by noise. To make up for it, \cite{li2019certified} propose stability training with noise (STN) and \cite{cohen2019certified} adopt training with noise, both of which enforce classifiers to learn the mapping between noisy inputs and the correct labels. However, there is no guarantee to ensure $g(x_0)$ to be correctly labeled. Actually we found the robustness mainly comes from the STN/training with noise, rather than the noise addition at the inference. In Fig.~\ref{fig:certify}(a), we observe that the classifiers trained without additive noise (triangle) degrades significantly compared with STN/training with noise (diamond/circle). The result is an evidence that a classifier almost cannot defend adversarial attacks when trained without but tested with additive noise (solid line with triangles). Therefore, we conclude the smoothed classifier can only improve robustness only if the base classifier is robust.

We consider robustness refers to the ability of a {\em DNN} to classify adversarial examples into the correct classes, and such an ability should be evaluated on the population of adversarial examples, not a single instance. 

%% file: experiment.tex
\section{Experiment}

\begin{table}[]\tiny \vspace{0mm}
	\centering
	\begin{tabular}{|c|cccc|}
		\hline
		\multicolumn{1}{|c|}{\bf Dataset}  &\multicolumn{1}{c}{\bf $\eta_{1}$} &\multicolumn{1}{c}{\bf $ \eta_2$}   &\multicolumn{1}{c}{\bf $ \gamma$} &\multicolumn{1}{c|}{\bf $\varepsilon$} 
		\\ \hline 
		MNIST        &$0.5/\gamma$ &$1\times10^{-4} $&$\{0.25, 1.5, 3\}$ &$\{0.84, 0.34, 0.21\}$ \\
		CIFAR-10 (R)     &$0.5/\gamma$ &$1\times10^{-4}$   &$\{0.25, 1.5, 5\}$ &$\{1.53,0.92,0.40\}$\\
		CIFAR-10 (V)     &$0.5/\gamma$ &$1\times10^{-4}$   & $\{0.25, 1.5, 5\}$ &$\{1.23,0.57,0.28\}$\\
		Tiny ImageNet  &$0.5/\gamma$ &$2\times10^{-5}$  &$1.5$ &$0.93$\\
		\hline
	\end{tabular} 
	\caption{Hyperparameters and perturbation ranges on different datasets. CIFAR-10 (R) represents CIFAR-10, ResNet-18 and CIFAR-10 (V) represents CIFAR-10, VGG-16.}
	\label{tab:NAL settings}
		\vspace{-5mm}
\end{table}

\textbf{Baselines, datasets and models.} Testing accuracies under different levels of adversarial attacks are chosen as the metric. We compare the empirical performance of NAL with representative baselines including: WRM (\cite{sinha2017certifying}), SmoothAdv (\cite{salman2019provably}), STN (\cite{li2019certified}), smoothing(\cite{cohen2019certified}), PGD(\cite{madry2017towards}) and TRADES (\cite{zhang2019theoretically}). Since WRM requires the loss function to be smooth, we follow the convention to adapt the ReLU activation layer to the ELU layer. 
TRADES is an adversarial training algorithm which won 1st place in the NeurIPS 2018 Adversarial Vision Challenge. Experiments are conducted on datasets MNIST, CIFAR-10, and Tiny ImageNet, and models including a three-layer CNN, ResNet-18, VGG-16, and their corresponding variants with ReLu replaced by ELU for fair comparison with WRM. The cross-entropy loss is chosen for $\ell$ and $ c(x,x_0) = \|x-x_0\|^2_2 $ is used as the cost function.

\begin{figure*}[t]\scriptsize
	\begin{minipage}[]{0.42\linewidth}
		\centering
		\includegraphics[width=0.9\linewidth]{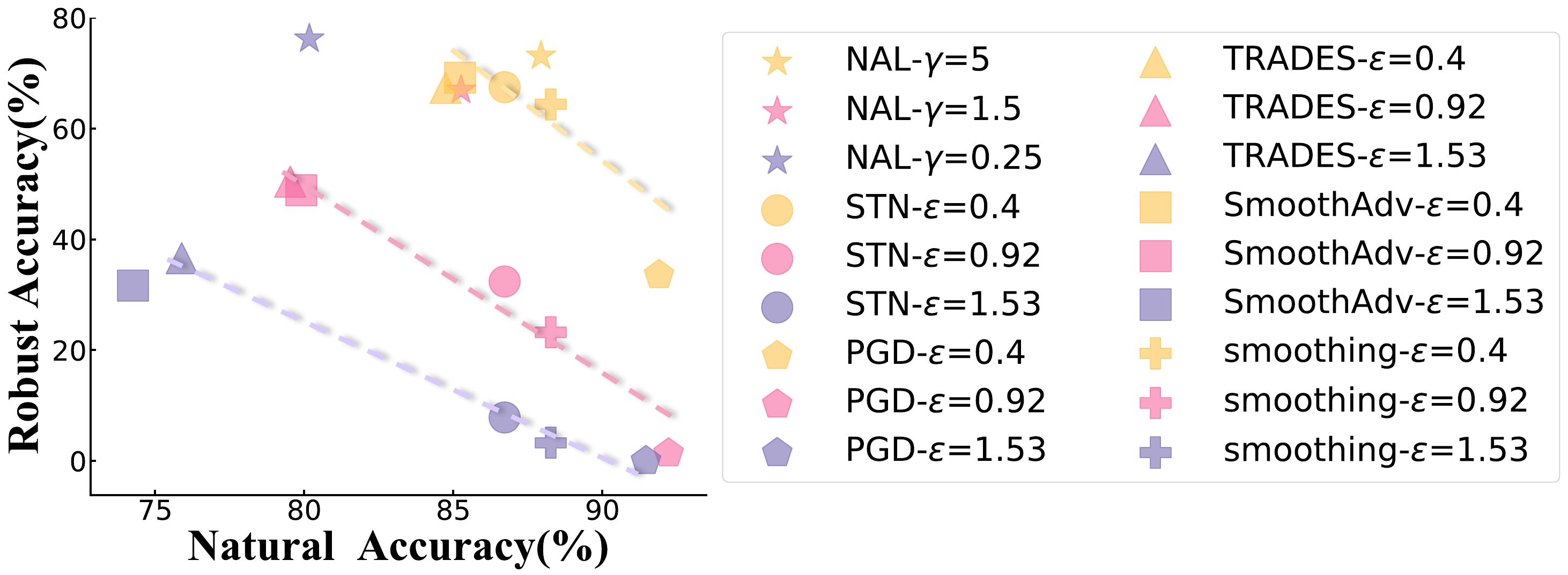}
		\caption{Trade-off between robustness and natural accuracy on CIFAR-10, ResNet-18 with $ \sigma = 0.1 $ and different $ \varepsilon $s. The dotted line shows the general trend of baselines with each $ \varepsilon $. The same color represents the same setting, under which NAL has the best performance overall. }
		\label{fig:scatter}
	\end{minipage}
	\begin{minipage}[]{0.58\linewidth}
		\centering
		\makeatletter\def\@captype{table}\makeatother	\caption{NAL outperforms baselines on MNIST (CNN), CIFAR-10 (ResNet-18), and Tiny ImageNet (ResNet-18) under PGD-20 atack with $ \gamma=1.5,  \varepsilon = 0.92.$ The best performance of each robust and natural accuracy are in bold.}
		\label{tab:emprical acc}
		\setlength{\tabcolsep}{0.7mm}
		\begin{tabular}{|c|c|c|c|c|c|c|}
			\hline
			& \multicolumn{2}{c|}{MNIST}          & \multicolumn{2}{c|}{CIFAR-10}       & \multicolumn{2}{c|}{TinyImageNet} \\ \hline
			Model     & Robust Accuracy  & Natural          & Robust Accuracy  & Natural          & Robust Accuracy      & Natural    \\ \hline
			PGD       & 97.14\%          & 99.04\%          & 0.88\%           & \textbf{91.90\%} & 4.66\%               & \textbf{63.66\%}    \\ \hline
			TRADES    & 98.14\%          & \textbf{99.19\%} & 50.43\%          & 79.53\%          & 34.90\%              & 56.72\%    \\ \hline
			STN       & 97.81\%          & 99.13\%          & 32.51\%          & 86.50\%          & 22.77\%              & 55.46\%    \\ \hline
			SmoothAdv & 98.26\%          & 99.16\%          & 48.71\%          & 80.06\%          & 33.00\%              & 56.64\%    \\ \hline
			Smoothing & 97.02\% 		&98.96\%  			& 22.99\%      		& 88.50\%			&20.63\%				&59.96\%			\\ \hline
			NAL       & \textbf{98.29\%} & 99.18\%          & \textbf{66.93\%} & 85.44\%          & \textbf{45.89\%}     & 59.63\%    \\ \hline
		\end{tabular}
	\end{minipage}
\vspace{-7mm}
\end{figure*}

\textbf{Training hyperparameters.} Table~\ref{tab:NAL settings} gives the training hyperparameters in NAL and the batch size is chosen as $128$. The hyperparameters used in baselines are supplied in supplementary file B.1.
Since NAL and WRM bound the adversarial perturbations by the Wasserstein distance $\rho$ which is different from the $\ell_{2}$-norm perturbation range $\varepsilon$ in other methods, we establish an equivalence between the perturbation ranges in different methods. Following the convention of \cite{sinha2017certifying}, we choose different $\gamma$s and for each $\gamma$ we generate adversarial examples $x$ by PGD with $15$ iterations. We compute $\rho$ as the expected transportation cost between the generated adversarial examples and the original inputs over the training set:
\begin{equation}\label{eq:fair_compare}
\varepsilon^{2} = \rho(\theta) = \mathbb{E}_{P_0}\mathbb{E}_{Z}\left[c\left( x+z, x_0\right)\right].
\end{equation}
And $\varepsilon$ can be computed accordingly. The corresponding values of $\gamma$ and $\varepsilon$ are given in Table~\ref{tab:NAL settings} as well.

\textbf{Metrics.} To evaluate the \textbf{certified accuracies} for the smoothed classifiers, we calculate the radius $ R $ for every point in the test set as mentioned in Sec.~\ref{sec: smoothed classifiers}, and compare $R$ with the attack radius to decide whether the point is robust. The same setting of \cite{cohen2019certified} is adopted to evaluate the certified accuracy.
We also evaluate the \textbf{empirical accuracies} for different methods by launching the PGD attack \cite{kurakin2016adversarial,madry2017towards} following the convention of \cite{li2019certified,sinha2017certifying,zhang2019theoretically}, etc. We set the number of iterations in PGD attack as $ K_{\text{PGD}} = 20, 100 $ respectively and the learning rate $ \eta = {2\varepsilon}/{K_{\text{PGD}}}$ where $ \varepsilon $ is $\ell_{2}$ attack radius. The result of $K_{\text{PGD}} =100 $ is included in the supplementary file B.4.

%% file: result.tex
\subsection{Results}
Due to space constraints, we only show partial results. Please find the complete results in supplementary file B.

\textbf{Noise level.} We vary the value of $\sigma$ in the experiments to find out their impact. By the results in Tab. \ref{tab:sigma change on cifar}, we observe $\sigma=0.1$ yields the best performance on CIFAR-10, considering all levels of adversarial attacks. Hence, we choose $\sigma=0.1$ by default in the following.

\begin{table}[htb]\scriptsize
	\centering
	\vspace{-3mm}
	\begin{tabular}{ll|lllll}
		\toprule
		\multicolumn{2}{c|}{$\ell_2$ attack radius} & 0      & 0.25   & 0.5    & 0.75   & 1         \\ \midrule
		NAL                 & $\sigma=0.05$           & 0.8579 & 0.7809 & 0.6761 & 0.5549 & 0.4262\\
		NAL                 & $\sigma=0.1$            & 0.8522 & \textbf{0.8155} & \textbf{0.7684} & \textbf{0.7140} & \textbf{0.6466} \\
		NAL                 & $\sigma=0.2$            & 0.8307 & 0.7781 & 0.7213 & 0.6498 & 0.5644  \\
		\midrule
		SmoothAdv           & $\sigma=0.05$           & 0.7643 & 0.7086 & 0.6378 & 0.5644 & 0.4841  \\
		SmoothAdv           & $\sigma=0.1$            & 0.8066 & 0.7264 & 0.6281 & 0.5376 & 0.4399  \\
		SmoothAdv           & $\sigma=0.2$            & 0.7411 & 0.6758 & 0.6079 & 0.5327 & 0.4689  \\
		\midrule
		STN                 & $\sigma=0.05$           & \textbf{0.8988} & 0.7347 & 0.4834 & 0.2594 & 0.1167 \\
		STN                 & $\sigma=0.1$            & 0.8669 & 0.7609 & 0.6164 & 0.4416 & 0.2847  \\
		STN                 & $\sigma=0.2$            & 0.8000 & 0.7060 & 0.5867 & 0.4695 & 0.3523  \\ 
		\bottomrule
	\end{tabular}
	\vspace{-2mm}
	\caption{Different methods with different levels of noise on CIFAR-10, ResNet-18, $ \gamma = 1.5 $ and $ (K,r) = (4,4) $. The best performance at the same noise level is in bold.}
	\label{tab:sigma change on cifar}
	\vspace{-2mm}
\end{table}

\textbf{Sample number and PGD iterations.} We also study the impact of the noise sample number $r$ and PGD iteration $K$ to the model robustness with CIFAR-10, ResNet-18 as an example. The results in Tab.~\ref{tab:changes with different s and K} show that while the model performance enhances with $K$, it does not necessarily increase with a larger $r$. For a combined consideration of computation overhead and accuracy, we choose $K=4, r=4$ by default, which is likely to deliver sufficiently good performance. 

\begin{table}[ht]\scriptsize\vspace{-3mm}
	\begin{tabular}{c|ccccc}
		\toprule
		\multicolumn{1}{c|}{$ \ell_2 $ attack radius} & \multicolumn{1}{c}{0} & \multicolumn{1}{c}{0.25} & \multicolumn{1}{c}{0.5} & \multicolumn{1}{c}{0.75} & \multicolumn{1}{c}{1} \\ \midrule
		$(K,r) = (4,1)$     & \textbf{ 0.8647 }               & 0.7540                   & 0.5950                  & 0.4297                   & 0.2814                    \\
		$(K,r) = (4,4)$      & 0.8546                & 0.7643                   & 0.6520                  & 0.5202                   & 0.3922                   \\
		$(K,r) = (4,8)$     & 0.8537                & 0.7622                   & 0.6482                  & 0.5171                   & 0.3846                   \\
		\midrule
		$(K,r) = (8,1)$     & 0.8593                & 0.7555                   & 0.6091                  & 0.4630                   & 0.3260                  \\
		$(K,r) = (8,4)$      & 0.8517                & \textbf{0.7663 }                  &  \textbf{0.6566 }                 & 0.5289                   & 0.3970           \\
		$(K,r) = (8,8)$    & 0.8493                & 0.7582                   & 0.6520                  &  \textbf{0.5302 }                  & \textbf{0.3978 }                      \\ \bottomrule
	\end{tabular}
\vspace{-2mm}
	\caption{Testing accuracies of NAL (CIFAR-10, ResNet-18) on a variety of $r$ and $K$. Under each setting, the model with the highest clean accuracy ($\ell_{2}$ attack radius = $0$) is chosen for testing. Numbers in bold represent the best performance in defending the attack.
	}
	\label{tab:changes with different s and K}\vspace{-2mm}
\end{table}

\textbf{Certificate.} To better understand how close the upper bound is to the true distributional risk, we plot our certificate $ \gamma \rho+\mathbb{E}_{\widehat{P}_{\text {test }}} \left[\phi_{\gamma}(\theta ; x_0)\right] $ against any level of robustness $\rho$, and the out-of-sample (test) worst-case performance $ \sup _{{S} \in {\mathcal{P}}} \mathbb{E}_{S} [\ell(\theta ; s)] $ for NAL (Fig. \ref{fig:certify}(b)). Since the worst-case loss is almost impossible to evaluate, we solve its Lagrangian relaxation for different values of $ \gamma $: for each chosen $ \gamma $, we compute the average distance to adversarial examples in the test set as $\widehat{\rho}_{\text {test }}(\theta):=\mathbb{E}_{\widehat{P}_{\text {test }}}\mathbb{E}_{Z}\left[c\left(x_{\star}+z, x_0\right)\right]$ where $\widehat{P}_{\text {test}}$ is the test data distribution and $ x_{\star} =\argmax_{x}\mathbb{E}_{Z}\left\{\ell(\theta ; x+z)-\gamma c(x+z, x_0)\right\} $ is the adversarial perturbation of $ x_0 $. The worst-case loss is given by $( \widehat{\rho}_{\text {test }}\left(\theta\right),  \mathbb{E}_{\widehat{P}_{\text {test }}}\mathbb{E}_{Z}\left[\ell\left(\theta ; x_{\star}+z\right)\right] )$. We also observe that, $ \widehat{\rho}_{\text {test }}(\theta) $ tends to increase with a higher noise level. Hence we need to keep the noise at an appropriate level to make our certificate tractable.

\textbf{ Certified accuracies} of NAL, SmoothAdv, Smoothing and STN on MNIST are given in Fig.~\ref{fig:certify}(c). Equivalent hyperparameters of $ \gamma=1.5  $ are set for all methods. We found that NAL has close performance to SmoothAdv but superior to others. Smoothing does not fit well to MNIST, as most $R$ values are the same, due to a lack of regularization terms. Results on TinyImgaeNet and CIFAR-10 are given in the supplementary file B.2.

\textbf{Empirical accuracies} of different methods are presented in Fig.~\ref{fig:certify} (d), Fig.~\ref{fig:scatter} and Tab.~\ref{tab:emprical acc}. For WRM-related comparison, all experiments are conducted on ELU-modified DNNs to ensure smoothness. NAL exceeds WRM for every $ \gamma $ over all the datasets including MNIST and CIFAR-10. Emprical accuracies of other baselines are presented in Tab.~\ref{tab:emprical acc}. NAL outperforms all others in robust accuracy while STN, Smoothing and PGD enjoys relatively higher natural accuracy. This may be explained by the inherent trade-off between natural accuracy and robustness (\cite{zhang2019theoretically}).  To show NAL indeed has a better tradeoff, we depict the robustness-accuracy with different $\varepsilon$s in Fig.~\ref{fig:scatter} on CIFAR10, ResNet-18. Most of the baselines exhibit some kind of tradeoff except that NAL has superior performance above all.



%% file: conclusion.tex
\section{Conclusion}
Our work views the robustness of a smoothed classifier from a different perspective, {\em i.e.,} the worst-case population loss over the input distribution. We provide a tractable upper bound (certificate) for the loss and devise a noisy adversarial learning approach to obtain a tight certificate. Compared with previous works, our certificate is practically meaningful and offers superior empirical robustness performance. 

%% file: appendix.tex
\section{Proofs}

\subsection{Proof of Theorem 2}\label{appendix:thm robust bound}
\begin{proof}
	We express the worst-case loss in its dual form with dual variable $\gamma$. By the weak dual property, we have
	\begin{equation}\label{eq:weak daul}
	\begin{split}
	&\sup _{P: W_c(P \oplus Z, P_0) \le \rho} \mathbb{E}_{P \oplus Z}  [\ell(\theta ; s)] \le \\
	&\inf _{\gamma \geq 0} \sup _{P: W_c(P \oplus Z, P_0) \le \rho} \left\{\mathbb{E}_{P \oplus Z} [\ell(\theta ; s)]-\gamma W_{c}(P \oplus Z, P_0)+\gamma \rho\right\},
	\end{split}
	\end{equation} 
	the left hand-side of which can be rewritten in integral form:
	\begin{equation}
	    \resizebox{.91\linewidth}{!}{$
		\displaystyle
	\begin{aligned}
	&\inf _{\gamma \geq 0} \sup _{P: W_c(P \oplus Z, P_0) \le \rho} \left\{\mathbb{E}_{P \oplus Z}  [\ell(\theta ; x+z)]-\gamma W_{c}(P \oplus Z, P_0)+\gamma \rho\right\}\\
	=&\inf _{\gamma \geq 0} \sup _{P: W_c(P \oplus Z, P_0) \le \rho} \left\{\int \ell(\theta ; x+z) d Z(z)P(x)-\gamma W_{c}(P \oplus Z, P_0)+\gamma \rho\right\}.
	\end{aligned}
	$}
	\end{equation}
	Note that for any $\pi \in \Pi(P \oplus Z, P_0)$, we have $\int f(s) d P \oplus Z=\iint f(s) d \pi(s, x_0) $. And by the definition of Wasserstein distance, we have
	\begin{equation} \label{eq:doubleint}
	    \resizebox{.91\linewidth}{!}{$
		\displaystyle
	\begin{split}
	&\inf _{\gamma \geq 0} \sup _{P: W_c(P \oplus Z, P_0) \le \rho} \left\{ \int \ell(\theta ; s) d P \oplus Z(s) -\gamma W_{c}(P \oplus Z, P_0) +\gamma \rho\right\}\\
	=&\inf _{\gamma \geq 0} \sup _{P: W_c(P \oplus Z, P_0) \le \rho} \left\{ \iint \ell(\theta ; s) d \pi(s, x_0) -\gamma \inf _{\pi \in \Pi(P \oplus Z, P_0)} \iint c(s, x_0) d \pi(s,  x_0) +\gamma \rho\right\}\\
	=&\inf _{\gamma \geq 0} \sup _{P: W_c(P \oplus Z, P_0) \le \rho} \left\{ \sup _{\pi \in \Pi(P \oplus Z, P_0)} \iint[\ell(\theta ; s)-\gamma c(s, x_0) ] d\pi(s, x_0)+\gamma \rho\right\}.
	\end{split}
	$}
	\end{equation}

	By the independence between $ z $ and $ x, x_0 ,$  one would obtain
	\begin{equation}
	\begin{split}
	&\iint[\ell(\theta ; s)-\gamma c(s, x_0) ] d\pi(s, x_0)\\
	=&\iiint[\ell(\theta ; x + z)-\gamma c(x + z, x_0) ]dZ(z) d\pi(x, x_0)
	\end{split}
	\end{equation}
	By taking the maximum over $x$,
	\begin{equation}
		\begin{split}
		&\iiint[\ell(\theta ; x + z)-\gamma c(x + z, x_0) ]dZ(z) d\pi(x, x_0) \\
		= &\iint \mathbb{E}_{Z}[\ell(\theta ; x + z)-\gamma c(x + z, x_0) ]d\pi(x, x_0) \\
		\leq &\iint\sup_{x} \left\{\mathbb{E}_{Z}[\ell(\theta ; x + z)-\gamma c(x + z, x_0) ]\right\} d\pi(x, x_0).
		\end{split}
	\end{equation}
	Fixing $x$ to be value that maximizes the expression to be integrated,  $ x $ in the formula is fixed, so we only need to integrate $ d\pi(x, x_0) $ on $ X $. So we can get:
	\begin{equation}
	\begin{split}
	&\iint\sup_{x} \left\{\mathbb{E}_{Z}[\ell(\theta ; x + z)-\gamma c(x + z, x_0) ]\right\} d\pi(x, x_0) \\
	= & \int_{x_0} \sup_{x} \left\{\mathbb{E}_{Z}[\ell(\theta ; x + z)-\gamma c(x + z, x_0) ]\right\} dP_0 (x_0) \\
	= & \mathbb{E}_{P_0}\sup_{x}\mathbb{E}_{Z}  [\ell(\theta ; x + z)-\gamma c(x + z, x_0) ]. \\
	\end{split}
	\end{equation}
	Because the distribution of $ z $ is definite and $ z $ is independent of $ x $, and supremum of $ P \oplus Z $ is replaced by the supremum of $ x $. Therefore, Eq.~\ref{eq:doubleint} can be written as
	\begin{equation}
	    \resizebox{.91\linewidth}{!}{$
		\displaystyle
		\begin{split}
		&\inf _{\gamma \geq 0} \sup _{P: W_c(P \oplus Z, P_0) \le \rho} \left\{ \sup _{\pi \in \Pi(P \oplus Z, P_0)} \iint[\ell(\theta ; s)-\gamma c(s, x_0) ] d\pi(s, x_0)+\gamma \rho\right\}\\
		\le &\inf _{\gamma \geq 0} \sup _{P: W_c(P \oplus Z, P_0) \le \rho} \left\{\sup _{\pi \in \Pi(P \oplus Z, P_0)} \mathbb{E}_{P_0}\sup_{x}\mathbb{E}_{Z}  [\ell(\theta ; x + z)-\gamma c(x + z, x_0) ] +\gamma \rho\right\} \\
		 = & \inf _{\gamma \geq 0} \left\{\mathbb{E}_{P_0}\sup_{x}\mathbb{E}_{Z}  [\ell(\theta ; x + z)-\gamma c(x + z, x_0) ]+\gamma \rho\right\}.
	\end{split}
	$}
	\end{equation}
	By plugging the above into Eq.~\ref{eq:weak daul}, we could get
	\begin{equation}\label{eq:last step in proof of thm1}
	\begin{split}
	&	\sup _{P: W_c(P \oplus Z, P_0) \le \rho} \mathbb{E}_{P \oplus Z} [\ell(\theta ; s)]\\ \le& \inf _{\gamma \geq 0} \left\{\mathbb{E}_{P_0}\sup_{x}\mathbb{E}_{Z}  [\ell(\theta ; x + z)-\gamma c(x + z, x_0) ]+\gamma \rho\right\}\\
	=&  \inf _{\gamma \geq 0} \left\{\mathbb{E}_{P_0}\left[\phi_{\gamma}(\theta ; x_0)\right]+\gamma \rho\right\} \le \mathbb{E}_{P_0}\left[\phi_{\gamma}(\theta ; x_0)\right]+\gamma \rho.
	\end{split}
	\end{equation}
	for any given $ \gamma \geq 0 $, which completes the proof.
\end{proof}

\subsection{Proof of Lemma 1}\label{appendix:smooth proof}

\begin{proof}
	The proof of $ \hat{\ell} $ being $ { \frac{2M}{\sigma^2} } $-smooth is equivalent to $ \nabla \hat{\ell} $ being $ { \frac{2M}{\sigma^2} } $-\textit{Lipschitz}. We apply the Taylor expansion in $ \nabla \hat{\ell} $ at $ x_0$ and set $ \delta = x_0-x $:
	\begin{equation}\label{eq:Taylor}
	\nabla \hat{\ell}(x_0) = \nabla \hat{\ell}(x) + \nabla^2 \hat{\ell}(x+\theta\delta) \delta ,
	\end{equation}
	where $ 0 <\theta<1 $. Hence we only need to prove $ \|\nabla^2 \hat{\ell}(x+\theta\delta)\|_2 $ is bounded since $\| \nabla \hat{\ell}(x+\delta) - \nabla \hat{\ell}(x) \|_2 = \| \nabla^2 \hat{\ell}(x+\theta\delta) \delta  \|_2$.
	By taking the first and second-order derivatives of $\hat{\ell}(x)$, we have
	\begin{equation}
	\nabla \hat{\ell}(x)  = \frac{1}{(2\pi)^{d/2}\sigma^{d+2}} \int_{\mathbb{R}^d} \ell(t) (t-x)\exp \left(-\frac{1}{2\sigma^2}\|x-t\|^{2}\right) d t,
	\end{equation}
	and 
	\begin{equation} \label{eq:second}
	\resizebox{.91\linewidth}{!}{$
		\displaystyle
		\begin{split}
		\nabla^2 \hat{\ell}(x)  = \frac{1}{(2\pi)^{d/2}\sigma^{d+2}} \int_{\mathbb{R}^d} \ell(t) \exp \left(-\frac{1}{2\sigma^{2}}\|x-t\|^{2}\right)[-\mI+\frac{1}{\sigma^{2}}(t-x)(t-x)^{\top}] d t. \\
		\end{split}
		$}
	\end{equation}
	We divide the right hand-side of Eq.~\ref{eq:second} into two halves with the first half:
	\begin{equation}
	\begin{split}
	&\|\frac{1}{(2\pi)^{d/2}\sigma^{d+2}} \int_{\mathbb{R}^d} \ell(t) \exp \left(-\frac{1}{2\sigma^{2}}\|x-t\|^{2}\right)(-\mI) dt\|_2\\
	= & \frac{1}{\sigma^{2}}\|\hat{\ell}(x)(-\mI)\|_2 \le \frac{1}{\sigma^{2}}\|\hat{\ell}(x)\|_2 \le \frac{M}{\sigma^2}.
	\end{split}
	\end{equation}
	The second half is 
	\begin{equation}
	\resizebox{.91\linewidth}{!}{$
		\displaystyle
		\begin{split}
		&\|\frac{1}{(2\pi)^{d/2}\sigma^{d+4}} \int_{\mathbb{R}^d} \ell(t) \exp \left(-\frac{1}{2\sigma^{2}}\|x-t\|^{2}\right)((t-x)(t-x)^{\top}) dt\|_2\\
		\le & \frac{1}{(2\pi)^{d/2}\sigma^{d+4}} \int_{\mathbb{R}^d} |\ell(t)| \exp \left(-\frac{1}{2\sigma^{2}}\|x-t\|^{2}\right)\|(t-x)(t-x)^{\top}\|_2 dt\\
		\le & \frac{M}{(2\pi)^{d/2}\sigma^{d+4}} \int_{\mathbb{R}^d} \exp \left(-\frac{1}{2\sigma^{2}}\|x-t\|^{2}\right)\|(t-x)(t-x)^{\top}\|_2 dt.
		\end{split}
		$}
	\end{equation}
	Due to the rank of the matrix $ (t-x)(t-x)^{\top} $ is $1$, its $ \ell_2 $ norm is easy to compute:
	\begin{equation}
	\| (t-x)(t-x)^{\top} \|_2 = (t-x)^{\top} (t-x).
	\end{equation} 
	Hence 
	\begin{equation}
	\resizebox{.91\linewidth}{!}{$
		\displaystyle
		\frac{M}{(2\pi)^{d/2}\sigma^{d+4}} \int_{\mathbb{R}^d} \exp \left(-\frac{1}{2\sigma^{2}}\|x-t\|^{2}\right)\|(t-x)(t-x)^{\top}\|_2 dt = \frac{M}{\sigma^{2}}.
		$}
	\end{equation}
	Finally, combining the two halves we get
	\begin{equation}
	\|\nabla^2 \hat{\ell}(x+\theta\delta)\|_2 \le \frac{2M}{\sigma^{2}}.
	\end{equation}
\end{proof}

\subsection{Proof of Corollary \ref{corollary:concave}}\label{appendix:proof of coeollary}
\begin{corollary}\label{corollary:concave}
	For any $ c: \mathcal{X}\times\mathcal{X} \rightarrow \mathbb{R}_{+} \cup\{\infty\}$ $1$-strongly convex in its first argument, and $\hat{\ell}: x \mapsto \mathbb{E}_{Z} [\ell(\theta ; x+z)] $ being $  \frac{2M}{\sigma^2}  $-smooth, the function $
	\mathbb{E}_{Z} \left\{\ell(\theta ; x+z)-\gamma c\left(x+z, x_{0}\right)\right\}$ is strongly concave in $x$ for any $ \gamma \geq  \frac{2M}{\sigma^2} $.
\end{corollary}
The proof is in Appendix \ref{appendix:proof of coeollary}. 
Note that here we specify the requirement on the transportation cost $c$ to be $1$-strongly convex in its first argument. The $ \ell_{2} $-norm cost satisfies the condition. Before showing how the strong concavity plays a part in the convergence, we illustrate our algorithm first.
\begin{proof}
	Since $ \hat{\ell} $ is $ { \frac{2M}{\sigma^2} } $-smooth and $ c $ is $1$-strongly convex in its first argument, we have 
	\begin{equation}
	\nabla^2_x\hat{\ell}(\theta ; x) \preceq { \frac{2M}{\sigma^2} } \mI, ~~\text{and}
	\end{equation}
	\begin{equation}
	\nabla^2_x \mathbb{E}_{Z} c(x+z,z_0) = \nabla^2_x \left[c(x,x_0)+d\sigma^2\right] = \nabla^2_x c(x,x_0) \succeq \mI.
	\end{equation}
	Therefore we have
	\begin{equation}
	\nabla^2_x \mathbb{E}_{Z} \left\{\ell(\theta ; x+z)-\gamma c\left(x+z, x_{0}\right)\right\} \preceq ({ \frac{2M}{\sigma^2} }-\gamma) \mI.
	\end{equation}
	Hence the strong concavity is proved for $ \gamma \geq {\frac{2M}{\sigma^2} } $.
\end{proof}

\subsection{Convergence Proof}\label{appendix:convergence}

We start with the required assumptions, which roughly quantify the robustness we provide.

\begin{assumption}\label{assumption B}
	The loss $\hat{\ell}: \Theta \times \mathcal{X} \rightarrow [0,M]$ satisfies the Lipschitzian smoothness conditions
	\begin{equation}
	    \resizebox{.91\linewidth}{!}{$
		\displaystyle
	\begin{array}{l}
	\|\nabla_{\theta} \hat{\ell}(\theta ; x)-\nabla_{\theta} \hat{\ell}(\theta^{\prime} ; x)\|_{*} \leq L_{\theta \theta}\|\theta-\theta^{\prime}\|,\|\nabla_{x} \hat{\ell}(\theta ; x)-\nabla_{x} \hat{\ell}(\theta ; x^{\prime})\|_{*} \leq L_{x x}\|x-x^{\prime}\|, \\
	\|\nabla_{\theta} \hat{\ell}(\theta ; x)-\nabla_{\theta} \hat{\ell}(\theta ; x^{\prime})\|_{*} \leq L_{\theta x}\|x-x^{\prime}\|,\|\nabla_{x} \hat{\ell}(\theta ; x)-\nabla_{x} \hat{\ell}(\theta^{\prime} ; x)\|_{*} \leq L_{x \theta}\|\theta-\theta^{\prime}\|.
	\end{array}
	$}
	\end{equation}
\end{assumption}
Let $\|\cdot\|_{*}$ be the dual norm to $\|\cdot\| ;$ we abuse notation by using the same norm $\|\cdot\|$ on $\Theta$ and $\mathcal{X}$. Here we have proved the second condition of Assumption \ref{assumption B} holds true by Theorem 1, with $ L_{x x} = { \frac{2M}{\sigma^2} } $. Therefore, if $ \hat{\ell} $ satisfies the other three conditions, we could adopt a similar proof procedure for Theorem 2 in \cite{sinha2017certifying} to prove the convergence of Algorithm 1.

\subsection{Proof of Corollary 1}\label{appendix:tighter bound}

\begin{proof}
	By Eq.~\ref{eq:last step in proof of thm1} we could get 
	\begin{equation}
	\sup _{P: W_c(P \oplus Z, P_0) \le \rho} \mathbb{E}_{P \oplus Z} [\ell(\theta ; s)] \le \inf_{\gamma \geq 0} \left\{\gamma \rho+\mathbb{E}_{P_0}\left[\phi_{\gamma}(\theta ; x_0)\right]\right\},
	\end{equation}
	where
	\begin{equation}
	\begin{split}
	\mathbb{E}_{P_0} \left[\phi_{\gamma}(\theta ; x_0)\right]
	= \mathbb{E}_{P_0} \left\{\sup _{x \in \mathcal{X}} \mathbb{E}_{Z} \left[\ell(\theta ; x+z)-\gamma c\left(x+z, x_{0}\right)\right]\right\}.\\
	\end{split}
	\end{equation}
	
	Since $\ell(\theta ; x+z)-\gamma c\left(x+z, x_{0}\right)$ is strongly concave for $x+z$ in \cite{sinha2017certifying}, by \textit{Jensen Inequality} we have for any fixed $x$,
	\begin{equation}
	\begin{split}
	&\mathbb{E}_{Z} \left\{\ell(\theta ; x+z)-\gamma c\left(x+z, x_{0}\right)\right\}\\
	\le &\ell(\theta ; \mathbb{E}_{Z}(x+z))-\gamma c\left(\mathbb{E}_{Z}(x+z), x_{0}\right)\\
	= &\ell(\theta ; x)-\gamma c\left(x, x_{0}\right).
	\end{split}
	\end{equation}
	Hence, the following inequality holds
	\begin{equation}
	\mathbb{E}_{P_0}\left[\phi_{\gamma}(\theta ; x_0)\right] \le
	\mathbb{E}_{P_0} \sup _{x \in \mathcal{X}} \left[\ell(\theta ; x)-\gamma c\left(x, x_{0}\right) \right].
	\end{equation}
	By Proposition 1 in \cite{sinha2017certifying}, we could get
	\begin{equation}
	\begin{split}
		&\inf_{\gamma \geq 0} \left\{\mathbb{E}_{P_0} \sup _{x \in \mathcal{X}} \left[\ell(\theta ; x)-\gamma c\left(x, x_{0}\right) \right] + \gamma \rho\right\}\\ 
	= &\sup _{P^{\prime}: W_{c}(P^{\prime}, P_0) \leq \rho} \mathbb{E}_{P^{\prime}}[\ell(\theta ; x)].
	\end{split}
	\end{equation}
	Finally, we can get Corollary 1 by concatenating the inequalities which completes the proof.
\end{proof}

\subsection{Connections between Robustness Certificates}\label{appendix:proposition}
\begin{proposition}\label{proposition:connect with smoothing}
	Let $ p_A, p_B $ denote the largest and second largest probabilities returned by the smoothed classifier $g(x_0)$ and $R=\frac{\sigma}{2}\left(\Phi^{-1}\left(\underline{p_{A}}\right)-\Phi^{-1}\left( \overline{p_{B}}\right)\right)$. We choose $ \ell $ as the cross-entropy loss in the smoothed loss function $ \hat{\ell}(x) =\mathbb{E}_{Z} [\ell(\theta ; x+z)],~z \sim Z = \mathcal{N}(0,\sigma^2I) $. If 
	\begin{equation}\label{eq:condition of prop}
	\hat{\ell}(\theta ; x) \le -\log\left(\Phi\left[\Phi^{-1}( \overline{p_{B}} )+\frac{\|x-x_0\|_2}{\sigma} \right]\right)
	\end{equation}
	holds, and the ground truth label $ y = c_{A} $, then $g(x)$ is robust against any $x$ such that $ \|x-x_0\|_2\le R $.
\end{proposition}
\begin{proof}
	By Theorem 1 of \cite{cohen2019certified}, we just need to prove the condition Eq.~\ref{eq:condition of prop} leads to the condition $ \|x-x_0\|_2 \le R  $. With $\ell$ being the cross-entropy loss,
	\begin{equation}
	\hat{\ell}(x) =\mathbb{E}_{Z} [\ell(\theta ; x+z)] = \mathbb{E}_{Z} [-\log(f^{(y)}(x+z))].
	\end{equation}
	Then we use \textit{Jensen Inequality} on $ -\log(x) $ to obtain
	\begin{equation}
	\hat{\ell}(x) =\mathbb{E}_{Z} [-\log(f^{(y)}(x+z))] \geq -\log[\mathbb{E}_{Z} f^{(y)}(x+z)].
	\end{equation}
	As $ y = c_A $, we have $ \mathbb{E}_{Z} f^{(y)}(x+z)  = P (f(x+z) = c_A) $. By Eq.~\ref{eq:condition of prop}, we have
	\begin{equation}
	\begin{split}
		-\log [P (f(x+z) = c_A)] \le \mathbb{E}_{Z} [-\log(f^{(y)}(x+z))] \\ \le  -\log\left(\Phi\left[\Phi^{-1}(\overline{p_{B}})+\frac{\|x-x_0\|_2}{\sigma} \right]\right).
	\end{split}
	\end{equation}
	And hence
	\begin{equation}
	P (f(x+z) = c_A) \geq \Phi\left[\Phi^{-1}(\overline{p_{B}})+\frac{\|x-x_0\|_2}{\sigma} \right].
	\end{equation}
	By the proof of Theorem 1 in \cite{cohen2019certified},
	\begin{equation}
	P (f(x+z) = c_A) =\Phi\left(\Phi^{-1}\left(\underline{p_{A}}\right)-\frac{\|x-x_0\|_2}{\sigma}\right),
	\end{equation}
	which leads to 
	\begin{equation}
	\Phi^{-1}\left(\underline{p_{A}}\right)-\frac{\|x-x_0\|_2}{\sigma} \geq \Phi^{-1}(\overline{p_{B}})+\frac{\|x-x_0\|_2}{\sigma}.
	\end{equation}
	Therefore,
	\begin{equation}
	\|x-x_0\|_2 \le \frac{\sigma}{2}\left(\Phi^{-1}\left(\underline{p_{A}}\right)-\Phi^{-1}\left(\overline{p_{B}}\right)\right) = R.
	\end{equation}
	To sum up, if Eq.~\ref{eq:condition of prop} holds and $x_0$ is correctly classified, $g(x)$ is robust within a $\ell_{2}$ ball with radius $R$. One can tell the loss on a single instance is weakly associated with the robustness of the model, and the condition of $g(x)$ being robust is quite stringent. It is not practical to sum up the single-instance loss to gauge the model robustness either.
\end{proof}

%% file: appendix_B.tex
\section{Experiments}

\subsection{Baseline Settings}\label{appendix:baseline setting}
We provide the training settings for baselines in Table \ref{tab:baseline settings}. The learning rate $ \eta_{1} $ is adjusted according to different $ \gamma $s and $ \varepsilon $s. The noise level ($\sigma$) is the same for all methods.

\begin{figure}[ht]
	\centering
	\subfigure[]{
		\begin{minipage}[t]{0.5\linewidth}
			\centering
			\includegraphics[width=1\linewidth]{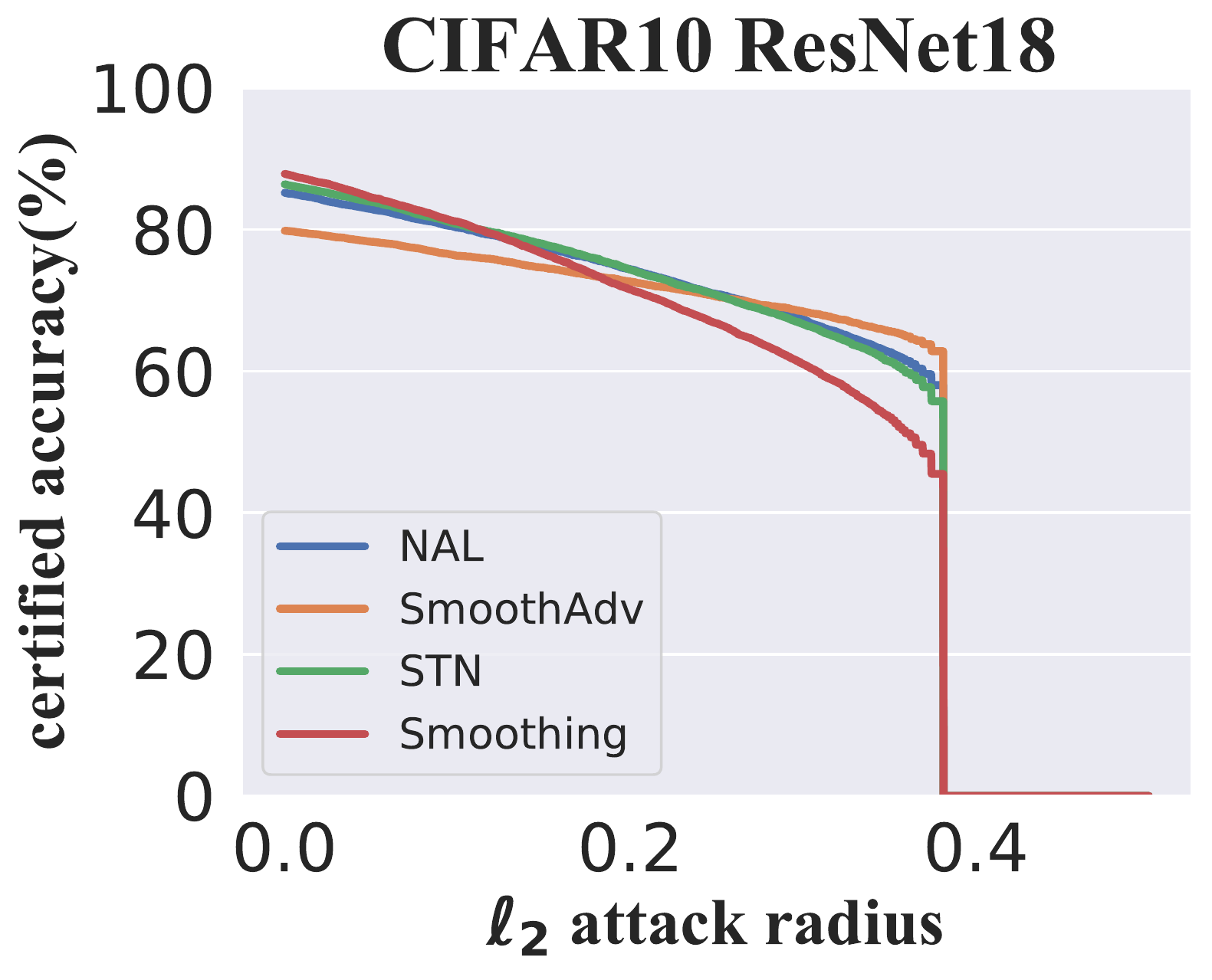}
		\end{minipage}
	}%
	\subfigure[]{
		\begin{minipage}[t]{0.5\linewidth}
			\centering
			\includegraphics[width=0.96\linewidth]{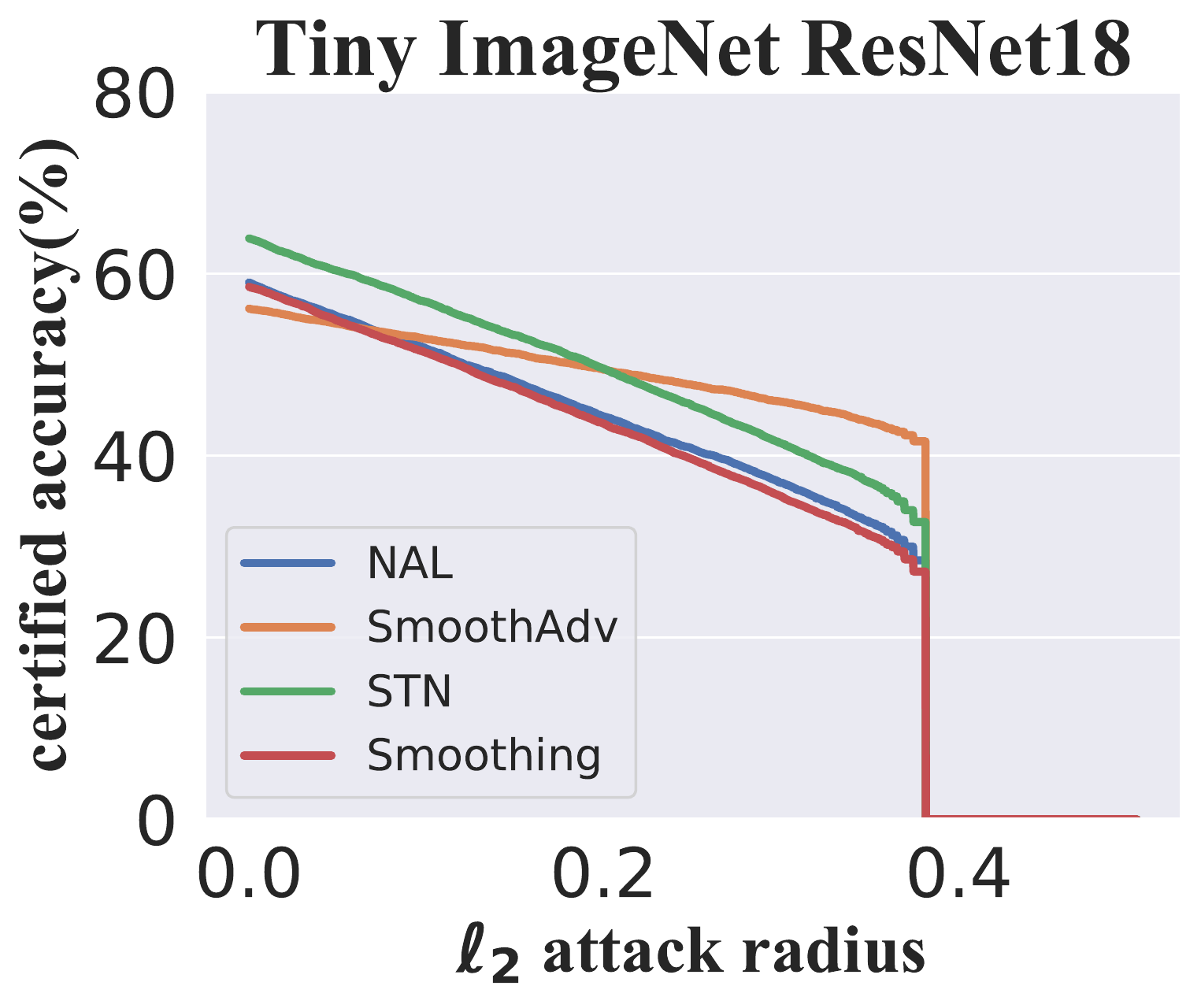}
		\end{minipage}%
	}%
	\caption{Compares NAL with SmoothAdv, Smoothing and STN on CIFAR-10 and Tiny ImageNet at $ \gamma=1.5 $ and the corresponding $ \varepsilon $.  }
	\label{fig:certify appendix}
\end{figure}

 \begin{table}\scriptsize
 	\centering
	\begin{tabular}{|c|lllll|}
		\hline
		\multicolumn{1}{|c|}{Dateset}                & mechanism & $\eta_1$ & $ \eta_2$ & \multicolumn{1}{c}{ batch size} & epochs     \\
		\hline
		\multicolumn{1}{|c|}{\multirow{4}{*}{MNIST}} & WRM       & $0.5/\gamma$         &$1\times10^{-4}$                              & 128                                                  &       25                                 \\
		\multicolumn{1}{|c|}{}                       & STN       & $-$         &$1\times10^{-4}$                              & 128                                                  &       25                                 \\
		\multicolumn{1}{|c|}{}                       & SmoothAdv  & $\epsilon/2$         &$1\times10^{-4}$                              & 128                                                  &       25                                \\
		\multicolumn{1}{|c|}{}                       & TRADES    & $\epsilon/2$          &$1\times10^{-4}$                              & 128                                                  &       25                                 \\
		
		\hline
		\multicolumn{1}{|c|}{\multirow{4}{*}{CIFAR-10}}                                   & WRM       & $0.5/\gamma$         &$1\times10^{-4}$                              & 128                                                  &       100                                 \\
		& STN       & $-$         &$1\times10^{-4}$                              & 128                                                  &      100                                 \\
		& SmoothAdv & $\epsilon/2$         &$1\times10^{-4}$                              & 128                                                 &       100                                \\
		& TRADES    & $\epsilon/2$          &$1\times10^{-4}$                              & 128                                                  &       100                                \\
		
		\hline
		\multicolumn{1}{|c|}{\multirow{4}{*}{Tiny ImageNet}}                        & WRM   & $0.5/\gamma$         &$2\times10^{-5}$                              & 128                                                  &       100                                \\
		& STN       & $-$         &$2\times10^{-5}$                              & 128                                                  &      100                                 \\ 
		& SmoothAdv  & $\epsilon/2$        &$2\times10^{-5}$                              & 128                                                  &       100                                 \\
		& TRADES     & $\epsilon/2$          &$2\times10^{-5}$                              & 128                                                  &       100                                  \\
		\hline
	\end{tabular}
	\caption{Baseline hyperparameter settings. $ \gamma $ and $ \varepsilon $ are chosen from the setting in experiments.}
	\label{tab:baseline settings}
\end{table}

\subsection{Certified Accuracy}
\begin{figure*}[ht]\vspace{0mm}
	\centering
	\subfigure[]{
		\begin{minipage}[t]{0.25\linewidth}
			\centering
			\includegraphics[width=1\linewidth]{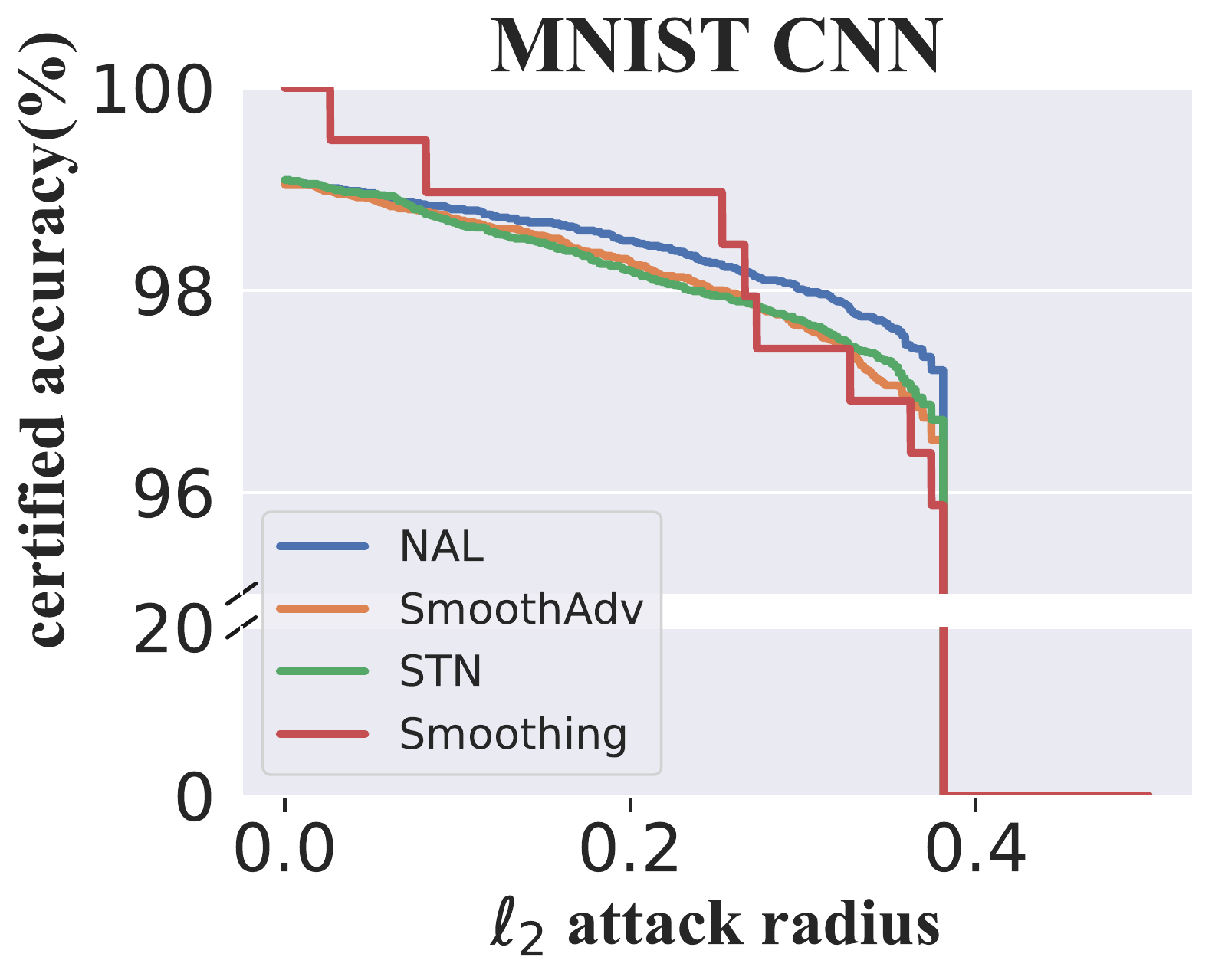}
		\end{minipage}%
	}%
	\subfigure[]{
		\begin{minipage}[t]{0.25\linewidth}
			\centering
			\includegraphics[width=1\linewidth]{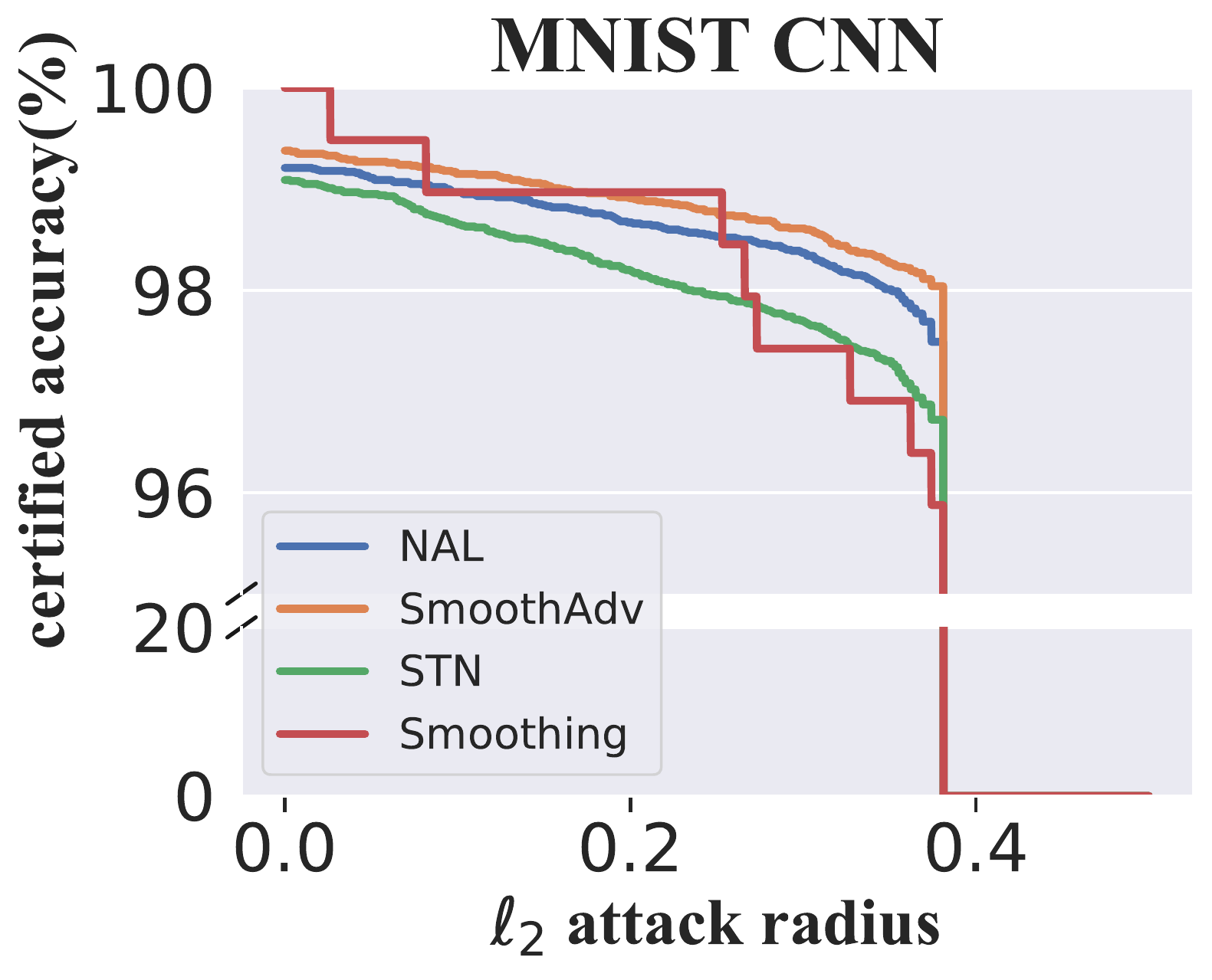}
		\end{minipage}%
	}%
	\subfigure[]{
		\begin{minipage}[t]{0.25\linewidth}
			\centering
			\includegraphics[width=1\linewidth]{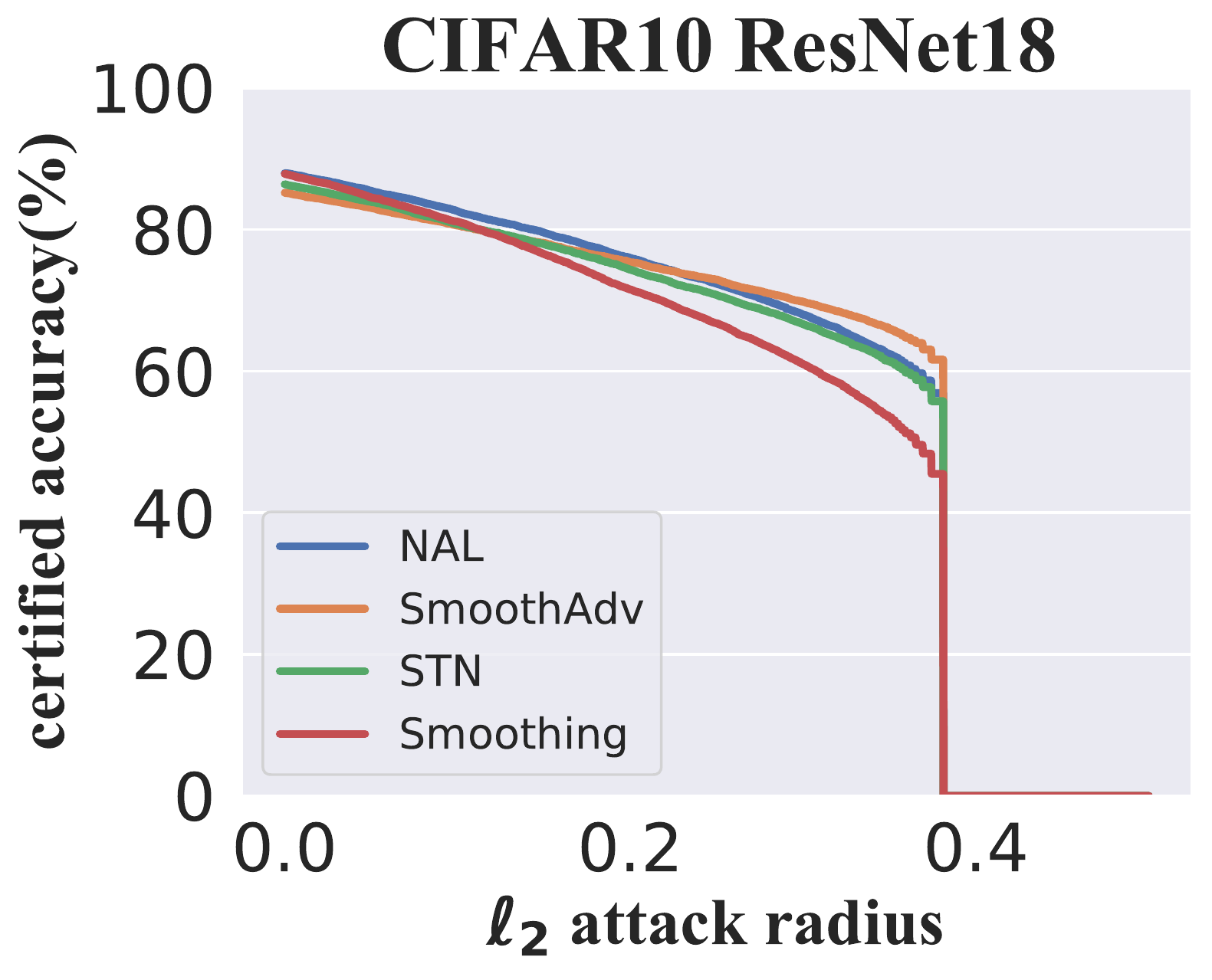}
		\end{minipage}%
	}%
	\subfigure[]{
		\begin{minipage}[t]{0.25\linewidth}
			\centering
			\includegraphics[width=1\linewidth]{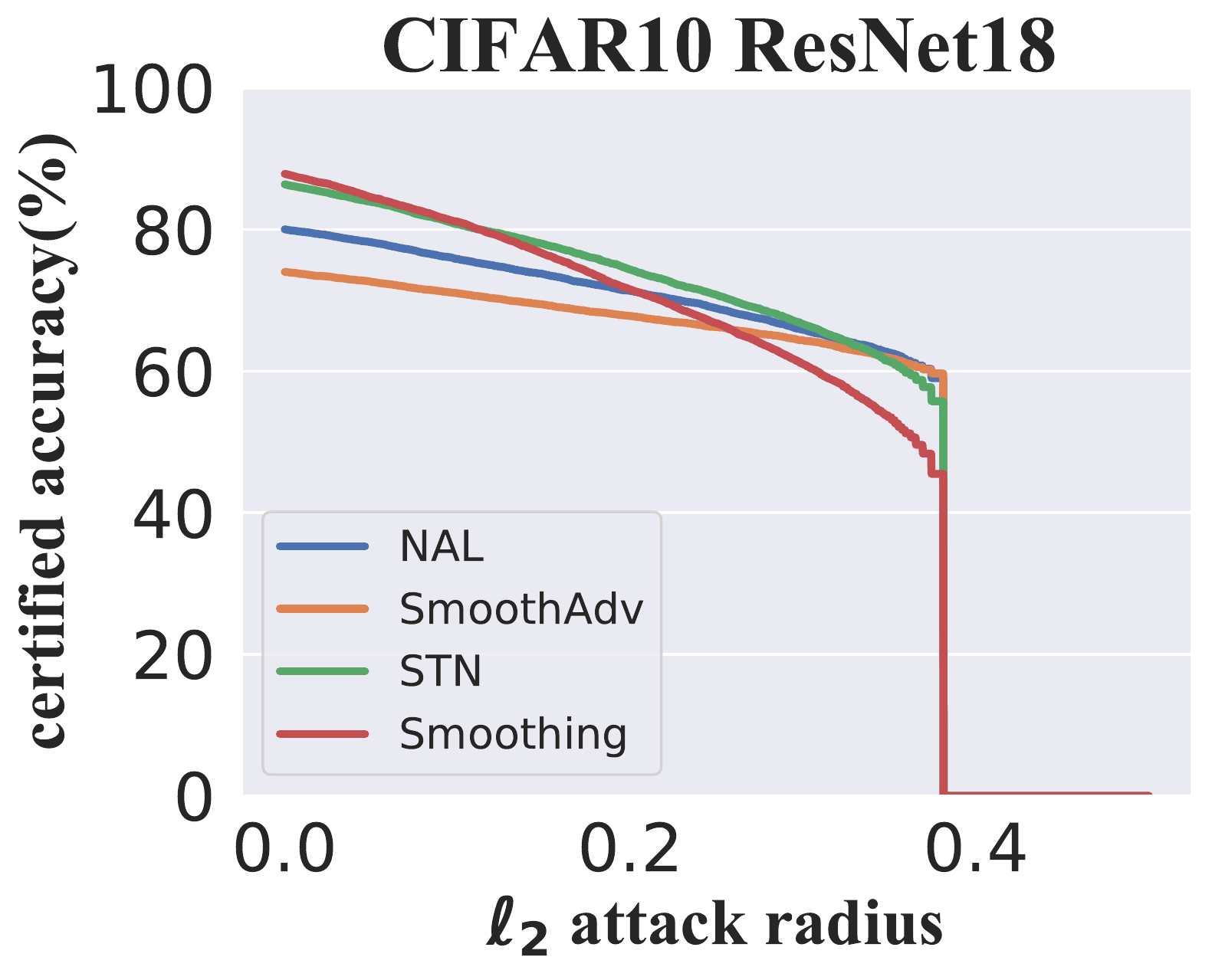}
		\end{minipage}%
	}%
	\vspace{-4mm}
	\caption{Certified accuracy on MNIST with $ \varepsilon = \{0.21, 0.84\} $, $ \gamma = \{3, 0.25\} $ corresponding to (a), (b) and CIFAR-10 with $ \varepsilon = \{0.4, 1.53\} $, $ \gamma = \{5, 0.25\} $ corresponding to (c), (d).  The smaller $ \gamma $ brings the better certified robustness on NAL. }
	\label{fig:certify acc2}
\end{figure*}

We present the result of certified accuracy on MNIST, CIFAR-10 and TinyImageNet in Fig.~\ref{fig:certify appendix} and Fig~\ref{fig:certify acc2}. In Fig.~\ref{fig:certify appendix}, NAL on TinyImgaeNet and CIFAR-10 performs approximately the same or slightly inferior to STN, mainly because there is no good reference for the selection of the parameter $ \gamma $, so using the same $ \gamma $ as MNIST may not be suitable for other datasets. And we present the certified accuracy on other $ \gamma $ on MNIST and CIFAR-10. We can observe that NAL has a better effect when the $ \gamma $ is relatively large (Fig,.~\ref{fig:certify acc2}). We believe that this is the effect of different parameter choices, and the maximum robustness radius will be affected by the number of samlping and the noise parameter $ \sigma $. 

\subsection{Empirical Accuracy } \label{appendix:parameter results}
We compare NAL with SmoothAdv and STN under the same experimental setting but different $ \sigma $s. In Table \ref {tab:sigma change on cifar}, NAL achieves the best performance at $\sigma = 0.1 $ above all. We believe in different experimental settings, the best $ \sigma $ value is different. For example, NAL and STN obtain the best performance at $ \sigma = 0.1 $, whereas SmoothAdv performs best at $\sigma = 0.05$. For the same $ \sigma $, NAL has superior performance than the other two baselines except that, when $ \sigma = 0.05 $, SmoothAdv is more robust than NAL for $ \ell_{2} $ attack radius $ \geq 0.75 $. This is mainly because SmoothAdv achieves the best performance when $ \sigma = 0.05 $. However, the model accuracies degrade below $0.5$ is not our main consideration.

\begin{figure*}[t]\vspace{0mm}
	\centering
	\subfigure[]{
		\begin{minipage}[t]{0.25\linewidth}
			\centering
			\includegraphics[width=1\linewidth]{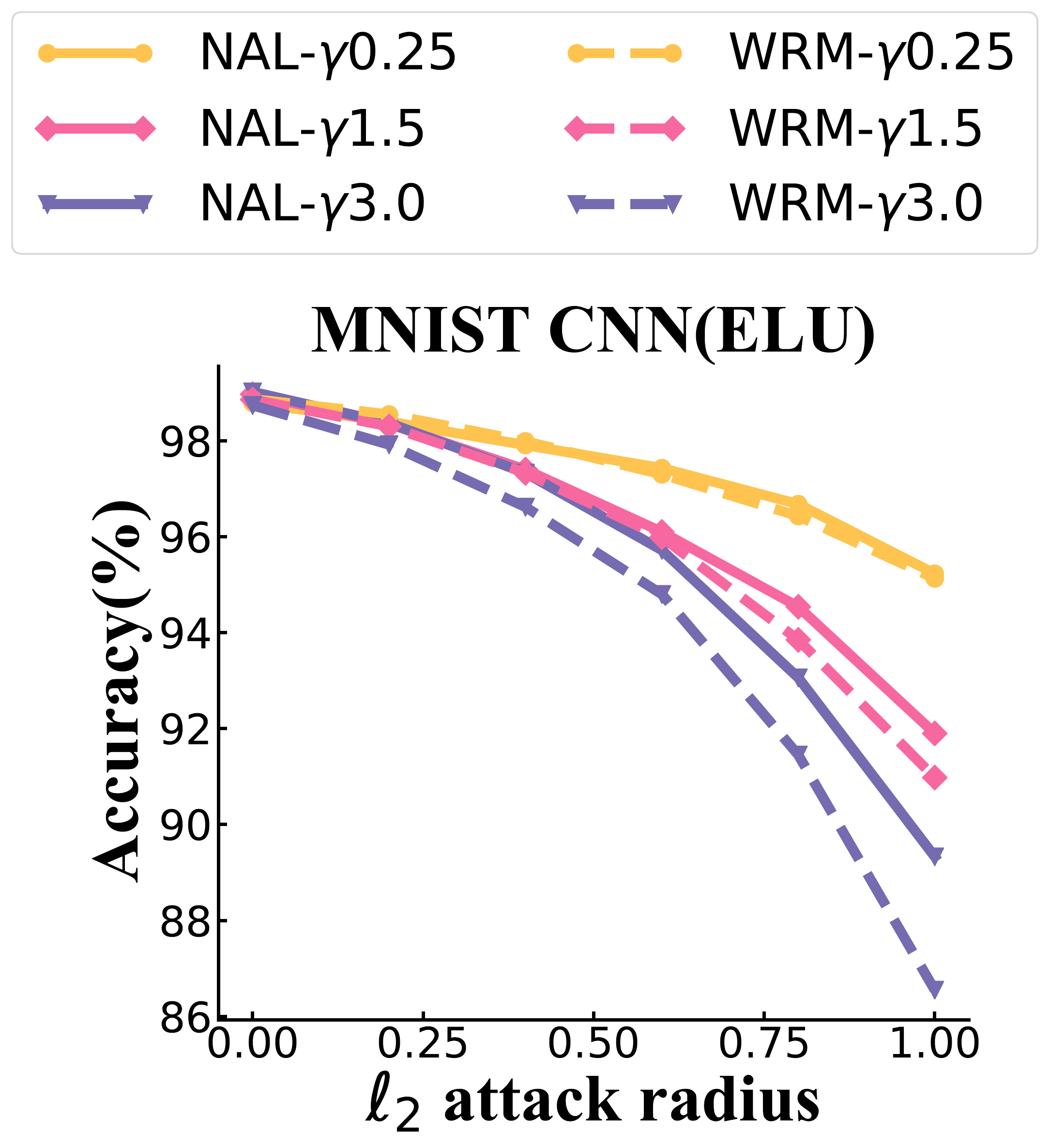}
		\end{minipage}%
	}%
	\subfigure[]{
		\begin{minipage}[t]{0.25\linewidth}
			\centering
			\includegraphics[width=1\linewidth]{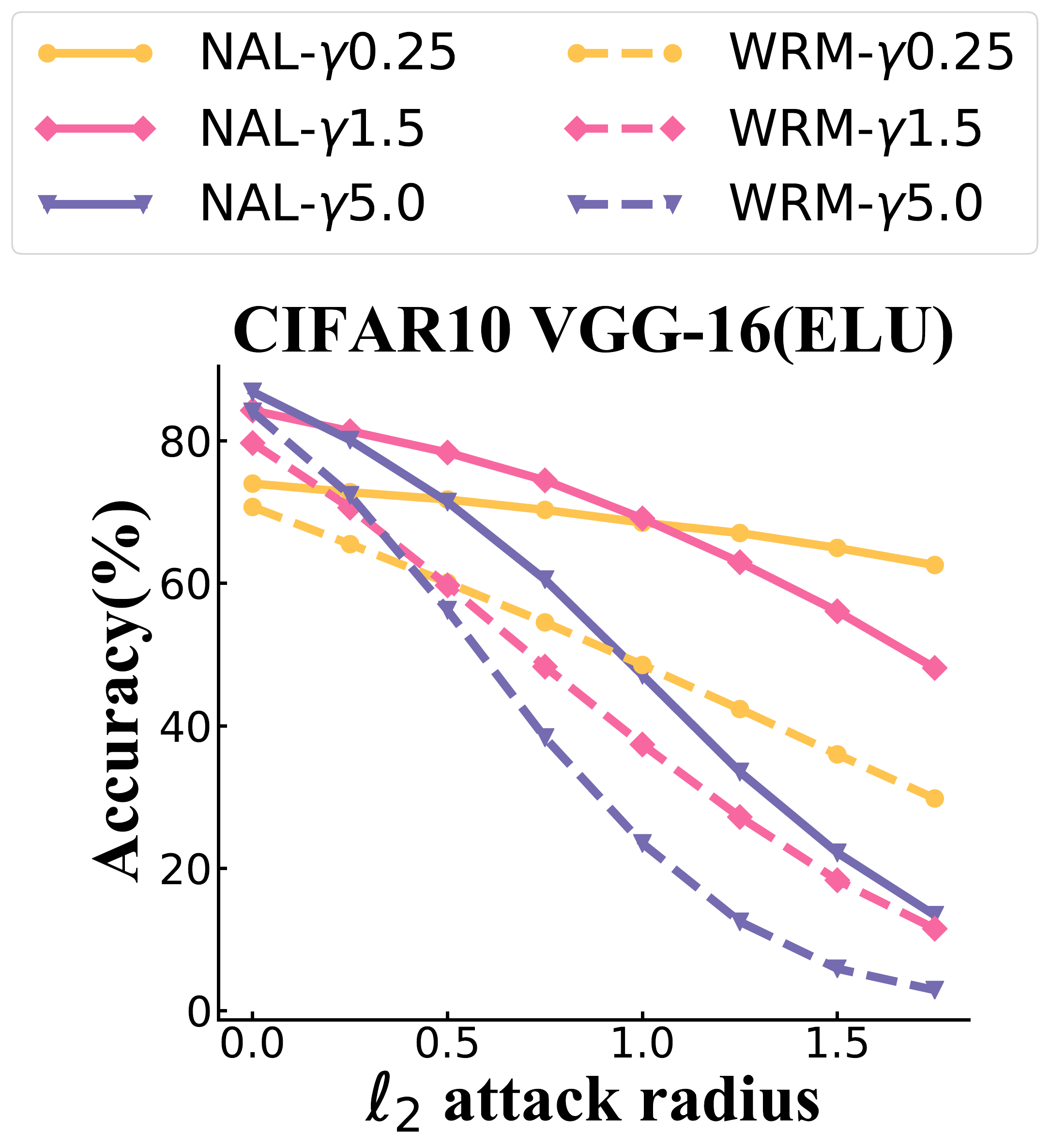}
		\end{minipage}%
	}%
	\subfigure[]{
		\begin{minipage}[t]{0.25\linewidth}
			\centering
			\includegraphics[width=1\linewidth]{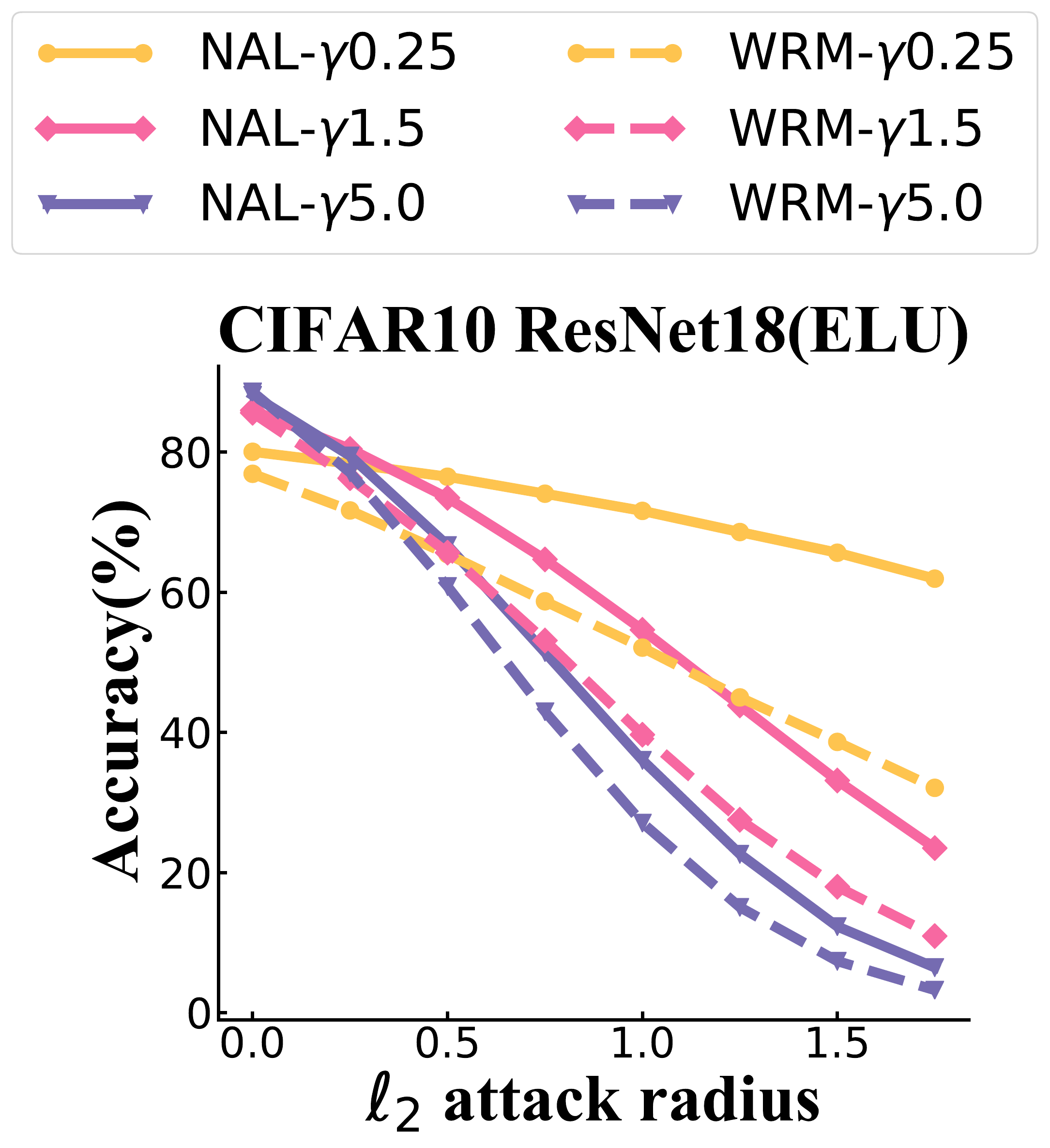}
		\end{minipage}%
	}%
	\subfigure[]{
		\begin{minipage}[t]{0.25\linewidth}
			\centering
			\includegraphics[width=1\linewidth]{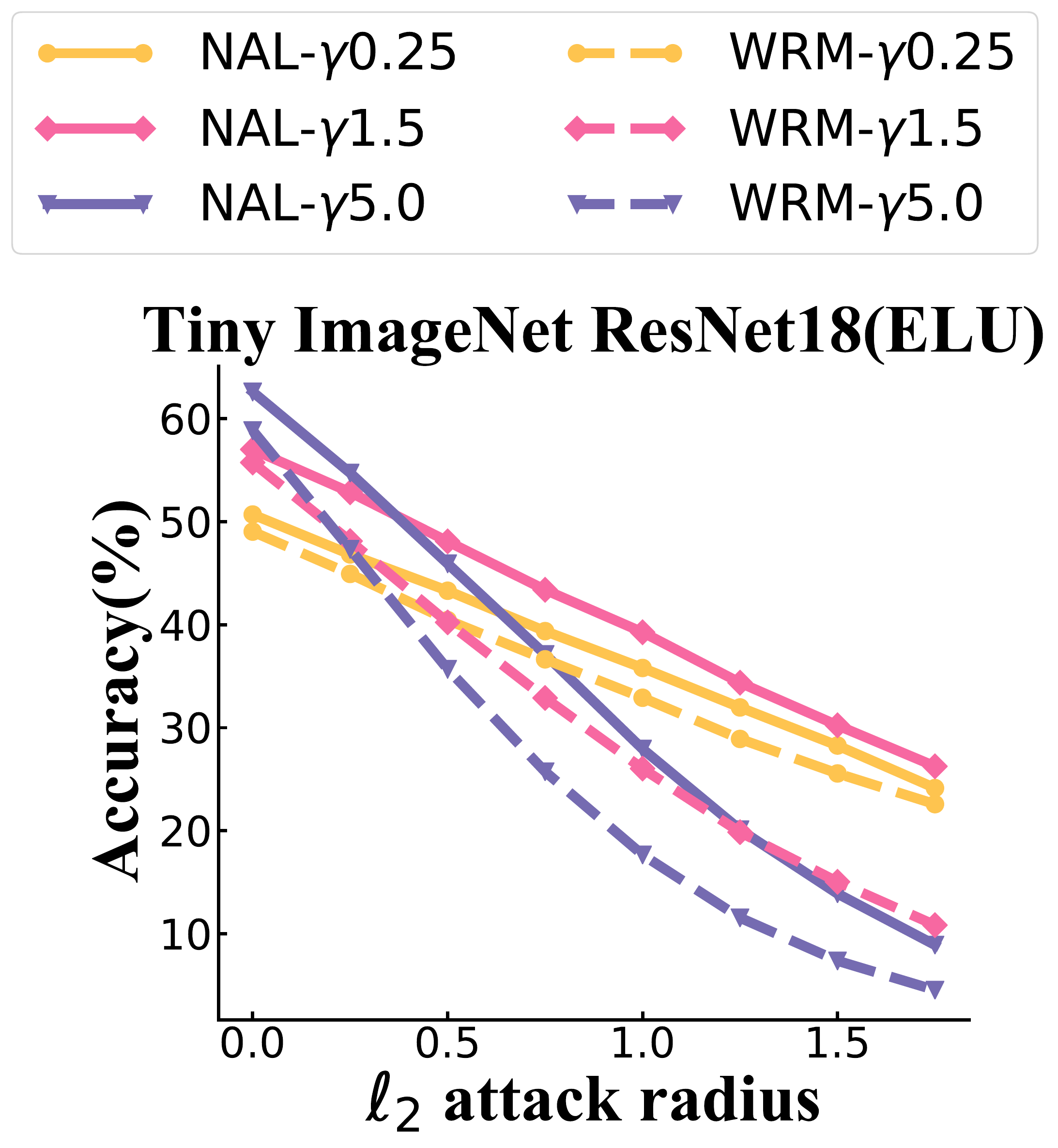}
		\end{minipage}%
	}%
	\vspace{-4mm}
	\caption{NAL outperforms WRM on MNIST, CNN, CIFAR-10, VGG-16 ,CIFAR-10, ResNet-18 and Tiny ImageNet, ResNet-18. All the experiments are trained on ELU models under different $ \gamma $s. For the same $\gamma$, NAL exceeds WRM.}
	\label{fig:ELU}
\end{figure*}

Fig. \ref{fig:ELU} shows the comparison between NAL and WRM on the ELU-based models under the PGD-20 attack. Table~\ref{tab:emprical acc on MINST and CIFAR with other eps} and \ref{tab:emprical acc pgd-100} show the comparison with SmoothAdv, TRADES, PGD, STN and Smoothing on regular models under PGD-20 and PGD-100 attack. NAL has superior robust performance than baselines in almost all cases except $ \gamma = 0.25, 3 $ over MNIST. However, in that case the robust accuracy of NAL is still the second highest and the gap between the robust accuracy of NAL and the best mechanism is smaller than $1\%$.

{\bf Results with varying $ \sigma $ and $ (K,r)$. }
In Table \ref{tab:the large table}, we show NAL's accuracy over a variety of $ \sigma, K, r$ values. We found that the result of $ \sigma=0.12 $ is generally better than a larger value. Under the same $ \sigma $, we choose $ K\in\{2,4,6,8\},r\in\{1,4\}, $. We found the model cannot converge with $ (K,r)=(2,1) $, and thus did not present the results. The results show that a larger $K$ admits better robustness whereas $r$ does not have much impact to the results.

{\bf Results of varying $\gamma$.}
We also show the impact of $\gamma$ on MNIST and CIFAR-10 in Table~\ref{tab:emprical acc on MINST and CIFAR with other eps}. On MNIST, $ \gamma $ takes the value $ \{0.25, 1.5, 3\} $ and $ \sigma $ is chosen as $0.05$. On CIFAR-10, $ \gamma \in \{0.25, 1.5, 5\} $ and $ \sigma $ is set to $0.1$. $(K,r) = (4,4) $ for all experiments. Fig. \ref{fig:ELU} compares NAL with WRM on models with ELU, and NAL exceeds WRM on every dataset for each $ \gamma $. 

\subsection{Comparison between ReLU and ELU}
\begin{figure}[H]
	\vspace{-0mm}
	\centering
	\includegraphics[width=0.7\linewidth]{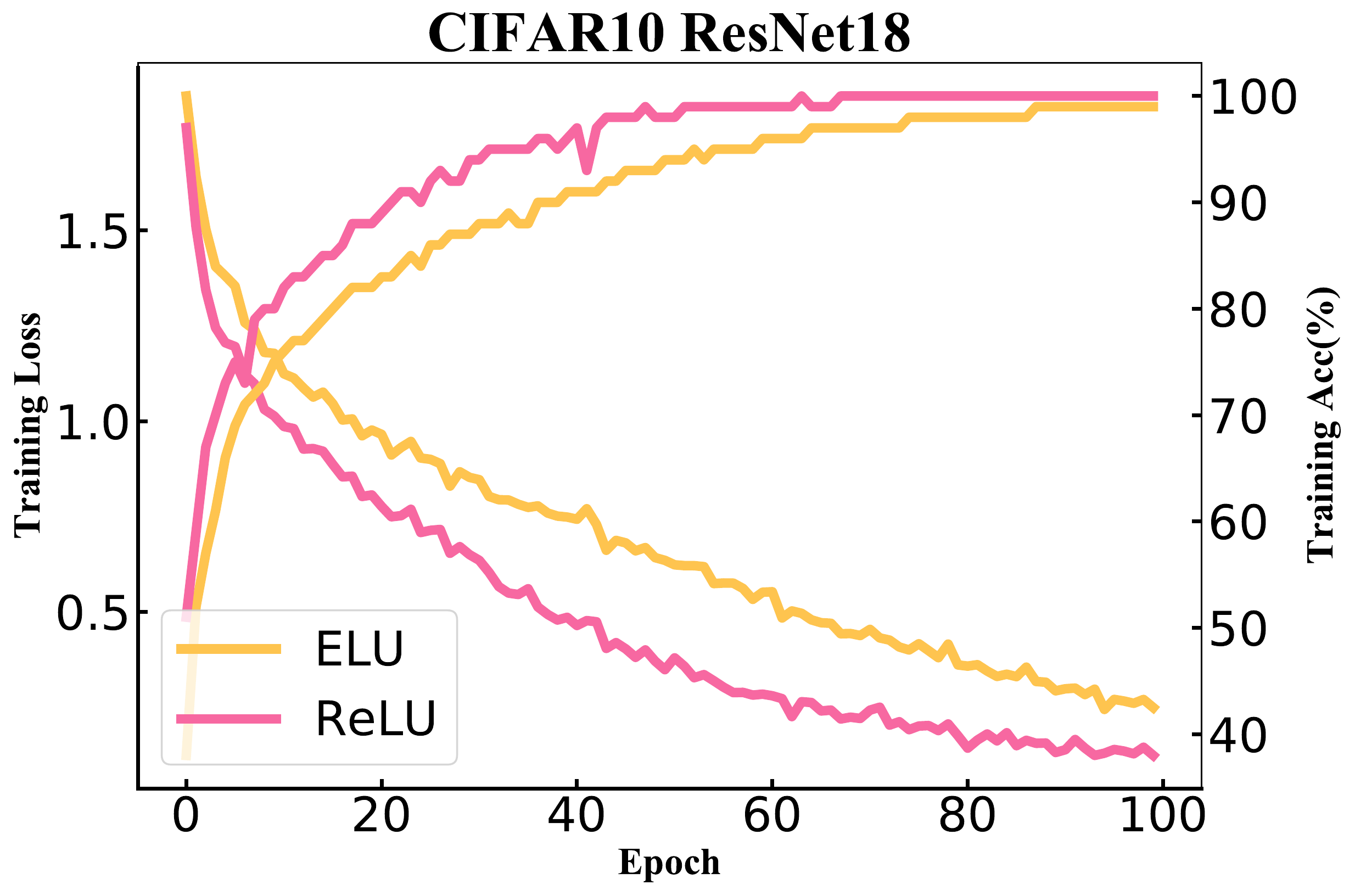}
	\caption{The comparison between the ReLU model (pink) and the ELU model (yellow) on CIFAR-10, ResNet-18 with $ \gamma = 1.5 $ and $ \sigma =0.1 $. The ReLU model converges faster than the ELU model.}
	\label{fig:cifareluvsrelu}
\end{figure}

Here we show the difference between ResNet-18 with ReLU and ELU on CIFAR-10 for NAL. From Fig. \ref{fig:cifareluvsrelu}, throughout the training process, the loss of the ReLU model is smaller than that of the ELU model, and ReLU model presents faster convergence. The robustness performance of both models is presented in Table \ref{tab:comparison between ReLU and ELU}. It is clear that in the testing phase, the ReLU model also obtains a better performance. Hence NAL generally yields better performance on ReLU models than ELU models.

\begin{table}[H]\tiny
	\centering
	\begin{tabular}{|c|ccccccc|}
		\hline
		$\ell_2$ attack radius & 0      & 0.25   & 0.5    & 0.75  & 1      & 1.25   & 1.5      \\ \hline
		ELU Model        & 0.8596 & 0.8046 & 0.7348 & 0.647 & 0.5465 & 0.4387 & 0.3315 \\
		ReLU Model       & 0.8522 & 0.8155 & 0.7684 & 0.714 & 0.6466 & 0.5684 & 0.4829  \\ \hline
	\end{tabular}
	\caption{Testing accuracies for the ReLU model and the ELU model on CIFAR-10, ResNet-18.  }
	\label{tab:comparison between ReLU and ELU}
\end{table}

\begin{figure}[H]
	\centering
	\includegraphics[width=0.7\linewidth]{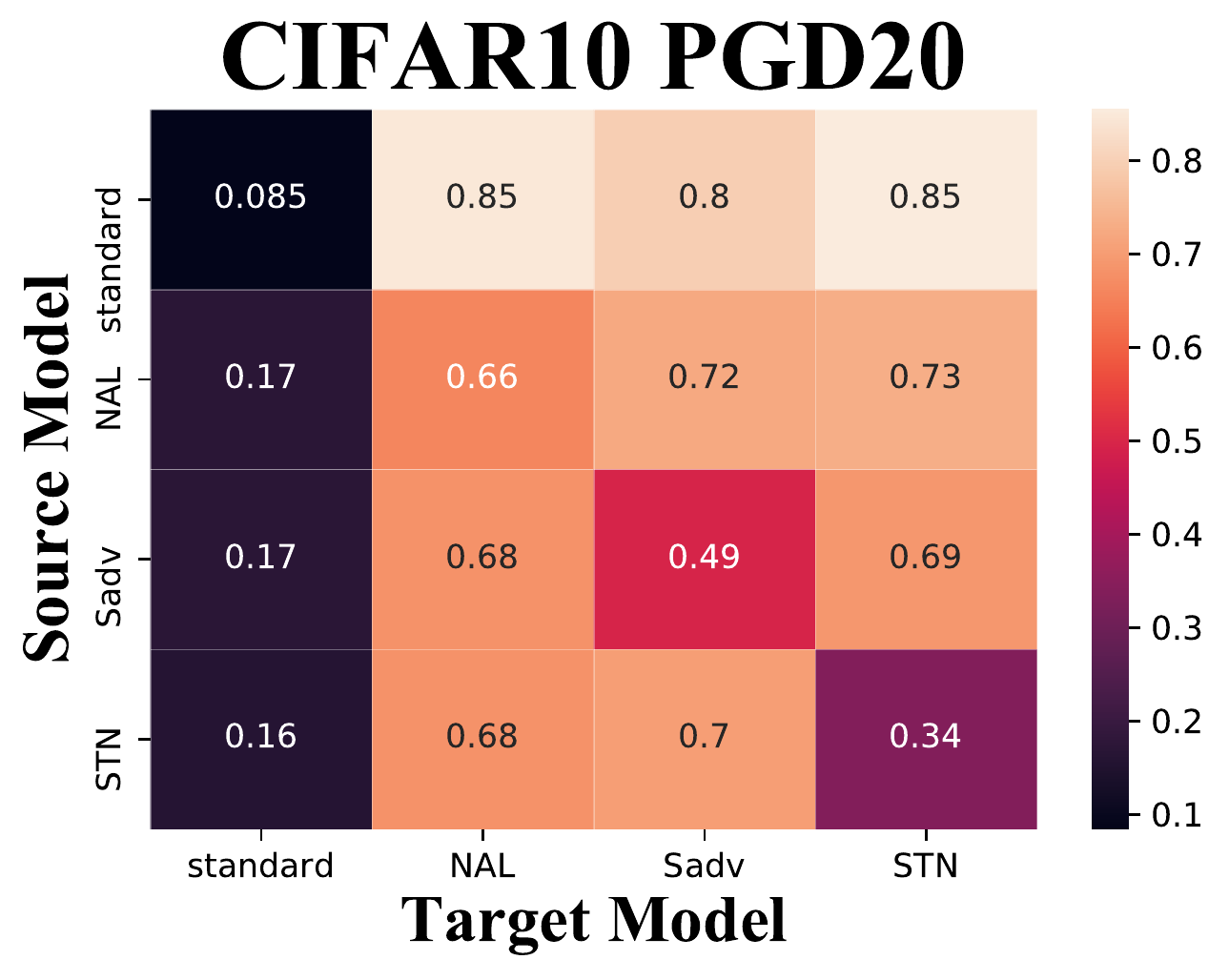}
	\caption{Testing accuracy of NAL and two baselines on CIFAR-10 under \textit{transfer-based black-box attacks}. The source model refers to the one used to craft adversarial examples, and the target model is the one being attacked.}
	\label{fig:heatmap}
\end{figure}

\subsection{Results with different attack source model}
Now we evaluate the robustness of the defenses on CIFAR-10 under black-box attacks to perform a thorough evaluation following the guidelines in \cite{carlini2019evaluating}. We evaluate \textit{transfer-based black-box attacks} using PGD-20. The results are presented in Fig.~\ref{fig:heatmap}. The vertical axis represents the source model where attack generated from, and the horizontal axis represents the target model which will be attacked. Obviously, the diagonal is the accuracy of the target model under the white-box attack, and the other places are the accuracy of the target model under the black-box attack. Obviously, NAL has better defense against white-box attacks and the accuracy under black-box attacks is also high. It also shows that these models obtain higher accuracy under transfer-based attacks than white-box attacks.

\begin{table*}[htb]
			\centering
	\begin{tabular}{ll|llllllll}
		\toprule
		\multicolumn{2}{c|}{$\ell_2$ attack radius} & 0      & 0.25   & 0.5    & 0.75   & 1      & 1.25   & 1.5    & 1.75   \\ \midrule
		NAL                 & $\sigma=0.05$           & 0.8579 & 0.7809 & 0.6761 & 0.5549 & 0.4262 & 0.2916 & 0.1888 & 0.1329 \\
		NAL                 & $\sigma=0.1$            & 0.8522 & \textbf{0.8155} & \textbf{0.7684} & \textbf{0.7140} & \textbf{0.6466} & \textbf{0.5684} & \textbf{0.4829} & \textbf{0.3909} \\
		NAL                 & $\sigma=0.2$            & 0.8307 & 0.7781 & 0.7213 & 0.6498 & 0.5644 & 0.4785 & 0.3837 & 0.2959 \\
		\midrule
		SmoothAdv           & $\sigma=0.05$           & 0.7643 & 0.7086 & 0.6378 & 0.5644 & 0.4841 & 0.4050 & 0.3297 & 0.2602 \\
		SmoothAdv           & $\sigma=0.1$            & 0.8066 & 0.7264 & 0.6281 & 0.5376 & 0.4399 & 0.3467 & 0.2700 & 0.2010 \\
		SmoothAdv           & $\sigma=0.2$            & 0.7411 & 0.6758 & 0.6079 & 0.5327 & 0.4689 & 0.4005 & 0.3350 & 0.2736 \\
		\midrule
		STN                 & $\sigma=0.05$           & \textbf{0.8988} & 0.7347 & 0.4834 & 0.2594 & 0.1167 & 0.0466 & 0.0155 & 0.0063 \\
		STN                 & $\sigma=0.1$            & 0.8669 & 0.7609 & 0.6164 & 0.4416 & 0.2847 & 0.1678 & 0.0927 & 0.0443 \\
		STN                 & $\sigma=0.2$            & 0.8000 & 0.7060 & 0.5867 & 0.4695 & 0.3523 & 0.2472 & 0.1708 & 0.1125 \\ 
		\bottomrule
	\end{tabular}
	\caption{Different methods with different levels of noise on CIFAR-10, ResNet-18, $ \gamma = 1.5 $ and $ (K,r) = (4,4) $. The best performance at the same noise level is in bold.}
	\label{tab:sigma change on cifar}
\end{table*}

\begin{table*}[hb]
	\centering
	\begin{tabular}{|l|ll|llllllll|}
		\hline
		\multicolumn{3}{|c|}{$\ell_2$ attack radius}                        & 0      & 0.25   & 0.5    & 0.75   & 1      & 1.25   & 1.5    & 1.75   \\ \hline
		\multicolumn{1}{|c|}{\multirow{7}{*}{$\sigma$ = 0.12}} & $K$=2 & $r$=4 &\textbf{0.8593} & 0.8414 & 0.8152 & 0.7891 & 0.7584 & 0.7177 & 0.6699 & 0.6103 \\
		\multicolumn{1}{|c|}{}                              & $K$=4 & $r$=1 & 0.8480 & 0.8142 & 0.7772 & 0.7306 & 0.6756 & 0.6185 & 0.5521 & 0.4748 \\
		\multicolumn{1}{|c|}{}                              & $K$=4 & $r$=4 & 0.8462 & 0.8129 & 0.7728 & 0.7237 & 0.6692 & 0.6044 & 0.5339 & 0.4646 \\
		\multicolumn{1}{|c|}{}                              & $K$=6 & $r$=1 & 0.8528 & 0.8312 & 0.8105 & 0.7803 & 0.7473 & 0.7081 & 0.6585 & 0.5989 \\
		\multicolumn{1}{|c|}{}                              & $K$=6 & $r$=4 & 0.8424 & 0.7990 & 0.7584 & 0.7022 & 0.6372 & 0.5663 & 0.4853 & 0.3950 \\
		\multicolumn{1}{|c|}{}                              & $K$=8 & $r$=1 & 0.8526 &\textbf{ 0.8418 }& \textbf{0.8329} & \textbf{0.8194} & \textbf{0.8049} & \textbf{0.7883} & \textbf{0.7670} & \textbf{0.7365} \\
		\multicolumn{1}{|c|}{}                              & $K$=8 & $r$=4 & 0.8443 & 0.8025 & 0.7494 & 0.6929 & 0.6250 & 0.5469 & 0.4578 & 0.3758 \\ \hline
		\multirow{8}{*}{$\sigma$ = 0.25}                       & $K$=2 & $r$=1 & 0.7522 & 0.6885 & 0.6186 & 0.5403 & 0.4596 & 0.3739 & 0.2971 & 0.2195 \\
		& $K$=2 & $r$=4 & 0.7953 & 0.7421 & 0.6801 & 0.6069 & 0.5226 & 0.4332 & 0.3498 & 0.2691 \\
		& $K$=4 & $r$=1 & 0.7669 & 0.7077 & 0.6399 & 0.5717 & 0.4896 & 0.4103 & 0.3318 & 0.2594 \\
		& $K$=4 & $r$=4 & 0.8121 & 0.7679 & \textbf{0.7143} & 0.6540 & 0.5882 & 0.5153 & 0.4381 & \textbf{0.3622 }\\
		& $K$=6 & $r$=1 & 0.7632 & 0.7082 & 0.6501 & 0.5790 & 0.5095 & 0.4349 & 0.3569 & 0.2805 \\
		& $K$=6 & $r$=4 & 0.8098 & 0.7578 & 0.7059 & 0.6410 & 0.5677 & 0.4875 & 0.4133 & 0.3300 \\
		& $K$=8 & $r$=1 & 0.7808 & 0.7285 & 0.6720 & 0.6073 & 0.5273 & 0.4490 & 0.3694 & 0.2951 \\
		& $K$=8 & $r$=4 & \textbf{0.8150} & \textbf{0.7694} & 0.7132 & \textbf{0.6549 }& \textbf{0.5896} & \textbf{0.5185} & \textbf{0.4391} & 0.3574 \\ \hline
		\multirow{8}{*}{$\sigma$ = 0.5}                        & $K$=2 & $r$=1 & 0.6434 & 0.5899 & 0.5335 & 0.4739 & 0.4150 & 0.3524 & 0.2958 & 0.2422 \\
		& $K$=2 & $r$=4 & 0.7122 & 0.6681 & 0.6201 & 0.5717 & 0.5212 & 0.4632 & 0.4090 & 0.3496 \\
		& $K$=4 & $r$=1 & 0.6744 & 0.6165 & 0.5631 & 0.5048 & 0.4450 & 0.3829 & 0.3226 & 0.2652 \\
		& $K$=4 & $r$=4 & 0.7186 & 0.6701 & 0.6217 & 0.5699 & 0.5135 & 0.4505 & 0.3966 & 0.3373 \\
		& $K$=6 & $r$=1 & 0.6860 & 0.6335 & 0.5799 & 0.5154 & 0.4566 & 0.3979 & 0.3362 & 0.2811 \\
		& $K$=6 & $r$=4 & 0.7185 & 0.6771 & 0.6243 & 0.5707 & 0.5174 & 0.4608 & 0.4019 & 0.3455 \\
		& $K$=8 & $r$=1 & 0.6943 & 0.6424 & 0.5911 & 0.5380 & 0.4758 & 0.4181 & 0.3548 & 0.2968 \\
		& $K$=8 & $r$=4 & \textbf{0.7239} & \textbf{0.6804} & \textbf{0.6320} & \textbf{0.5836} & \textbf{0.5345} & \textbf{0.4736} & \textbf{0.4250} & \textbf{0.3716} \\ \hline
	\end{tabular}
	\caption{NAL with different $ \sigma $s and $ (K,r) $ on CIFAR-10, ResNet-18 when $ \gamma = 1.16 $. The best performance under the same noise level is in bold.}
	\label{tab:the large table}
\end{table*}

\begin{table*}\footnotesize
	\begin{tabular}{|c|c|c|c|c|c|c|c|c|}
		\hline
		    & \multicolumn{2}{c|}{MNIST,~$ \varepsilon = 0.21 $} & \multicolumn{2}{c|}{MNIST,~$ \varepsilon = 0.84 $} & \multicolumn{2}{c|}{CIFAR-10,~$ \varepsilon = 0.4 $} & \multicolumn{2}{c|}{CIFAR-10,~$ \varepsilon = 1.53 $} \\ \hline
		Model     & Robust Accuracy  & Natural & Robust Accuracy  & Natural & Robust Accuracy   & Natural   & Robust Accuracy   & Natural   \\ \hline
		PGD       & 98.41\%          & 99.14\% & 85.51\%          & 98.85\% & 33.53\%           & \textbf{91.90\% }  & 0.00\%            & \textbf{91.46\%}   \\ \hline
		TRADES    & \textbf{98.65\% }         & \textbf{99.24\% }& 96.12\%          & 99.33\% & 67.43\%           & 84.74\%   & 36.66\%           & 75.89\%   \\ \hline
		STN       & 97.81\%          & 99.13\% & \textbf{97.81\%  }        & 99.13\% & 31.93\%           & 86.50\%   & 31.93\%           & 86.50\%   \\ \hline
		SmoothAdv & 98.45\%          & 99.05\% & 96.39\%          & \textbf{99.38\%} & 69.22\%           & 85.25\%   & 31.73\%           & 74.22\%   \\ \hline
		Smoothing    & 97.09\%          & 98.96\% & 97.09\%          & 98.96\% & 22.47\%           & 88.50\%   & 22.47\%           & 88.50\%   \\ \hline
		NAL       & 98.61\%          & 99.04\% & 97.18\%          & 99.29\% & \textbf{73.37\%}           & 88.14\%   & \textbf{76.19\%}           & 80.24\%   \\ \hline
	\end{tabular}
\caption{Empirical accuracy on MNIST with $ \varepsilon = \{0.21, 0.84\} $, $ \gamma = \{3, 0.25\} $ and CIFAR-10 with $ \varepsilon = \{0.4, 1.53\} $, $ \gamma = \{5, 0.25\} $ under PGD-20 attack. NAL outperforms baselines on CIFAR-10. And the gap between the robust accuracy of NAL and the best mechanism is smaller than 1\%.  The best performance of each robust and natural accuracy are in bold.}
\label{tab:emprical acc on MINST and CIFAR with other eps}
\end{table*}

\begin{table*}[]
	\centering
	\begin{tabular}{|c|c|c|c|c|c|c|}
		\hline
		PGD-100    & \multicolumn{2}{c|}{MNIST}          & \multicolumn{2}{c|}{CIFAR-10}       & \multicolumn{2}{c|}{TinyImageNet}   \\ \hline
		Model     & Robust Accuracy  & Natural          & Robust Accuracy  & Natural          & Robust Accuracy  & Natural          \\ \hline
		PGD       & 97.15\%          & 99.04\%          & 0.84\%           & \textbf{91.90\%} & 4.67\%           & \textbf{63.66\%} \\ \hline
		TRADES    & 98.14\%          & \textbf{99.19\%} & 50.15\%          & 79.53\%          & 34.88\%          & 56.72\%          \\ \hline
		STN       & 97.81\%          & 99.13\%          & 31.93\%          & 86.50\%          & 22.68\%          & 55.46\%          \\ \hline
		SmoothAdv & 98.27\%          & 99.16\%          & 48.31\%          & 80.06\%          & 32.98\%          & 56.64\%          \\ \hline
		Smoothing    & 97.09\%          & 98.96\%          & 22.47\%          & 88.50\%          & 20.73\%          & 59.96\%          \\ \hline
		NAL       & \textbf{98.29\%} & 99.18\%          & \textbf{66.73\%} & 85.44\%          & \textbf{45.71\%} & 59.63\%          \\ \hline
	\end{tabular}
	\caption{NAL outperforms baselines on MNIST (CNN), CIFAR-10 (ResNet-18), and Tiny ImageNet (ResNet-18) under PGD-100 atack with $ \gamma=1.5,  \varepsilon = 0.92.$ The best performance of each robust and natural accuracy are in bold.}
	\label{tab:emprical acc pgd-100}
\end{table*}